\documentclass{article}

\usepackage{lineno,hyperref}
\modulolinenumbers[5]

\usepackage{caption}
\usepackage{url}
\usepackage{color}

\usepackage{amsfonts, amssymb, bbm}
\usepackage{mathtools}
\usepackage{thmtools, thm-restate}

\usepackage{multirow}

\usepackage{float}
\usepackage{graphics,graphicx,epsfig,epstopdf}
\usepackage{psfrag}
\usepackage{subfigure}
\usepackage{rotating}
\usepackage{tikz,pgfplots}
\usepackage{framed}

\usepackage{algorithm}
\usepackage[noend]{algpseudocode}

\DeclareMathOperator*{\argmin}{arg\,min}
\DeclareMathOperator*{\argmax}{arg\,max}

\newcommand{\Att}{$\mathcal{A}$}
\newcommand{\Def}{$\mathcal{D}$}
\newtheorem{theorem}{Theorem}
\newtheorem{proposition}{Proposition}
\newtheorem{lemma}{Lemma}
\newtheorem{observation}{Observation}
\newtheorem{definition}{Definition}

\begin{document}
	
\title{Facing Multiple Attacks in Adversarial Patrolling Games with Alarmed Targets}

\author{Giuseppe De Nittis and Nicola Gatti\\ \small{Politecnico di Milano}\\ \small{Piazza Leonardo da Vinci, 32}\\ \small{20133 Milano, Italy}\\ \small{\{giuseppe.denittis,nicola.gatti\}@polimi.it}}
\date{}

\maketitle
	
\vspace{-0.5cm}
	
\begin{abstract}
	We focus on adversarial patrolling games on an \emph{arbitrary} graph, in which the \textit{Defender} can control a mobile resource and the targets are alarmed by an alarm system, while the \textit{Attacker} can observe the movements of the mobile resource of the \textit{Defender}, and, exploiting multiple attacking resources, perform different sequential attacks against the targets. This scenario captures, e.g., the terroristic assaults in Paris in 2015.
	It can be modelled as a zero-sum extensive-form game in which each player can play multiple times. The game tree is exponentially large both in the size of the graph and in the number of resources available to the Attacker.
	We show that, when the number of the Attacker's resources is free, the problem of computing the equilibrium path is $\mathsf{NP}$-hard even if the Attacker is restricted to play all her resources simultaneously, while, when the number of resources is fixed, the equilibrium path can be computed in polynomial time. In particular, we provide a dynamic-programming algorithm that, given the number of the Attacker's resources, computes the equilibrium path requiring polynomial time in the size of the graph and exponential time in the number of the resources. 
	Furthermore, since in real-world scenarios it is implausible that the Defender perfectly knows the number of resources of the Attacker, we study the robustness of the Defender's strategy when she makes a wrong guess about that number.
	We show that even the error of just a single resource, either underestimating or overestimating the number of the Attacker's resources, can lead to an arbitrary inefficiency, when the inefficiency is defined, as usual, as the ratio of the Defender's utilities obtained with a wrong guess and a correct guess. 
	We argue that, in this case, a more suitable definition of inefficiency is given by the difference of the Defender's utilities. With this definition, we observe that the higher the error in the estimation, the higher the loss for the Defender.
	Then, we investigate the performance of \emph{online} algorithms when no information about the  Attacker's resources is available, showing that there are no competitive deterministic algorithms.
	Finally, we resort to randomized online algorithms showing that we can obtain a competitive factor that is twice better than the one that can be achieved by any deterministic online algorithm.
\end{abstract}

\section{Introduction}
In this paper, we study the first security game in which an Attacker can observe the movements of a patroller on an \emph{arbitrary} graph and perform different attacks, either simultaneous or sequential, exploiting multiple resources. 
Differently with respect to  security games customarily studied in the literature, here we model our problem as a 2-player extensive-form game, with each player able to play multiple times. 
The defending agent, hereafter called \textit{Defender}, has a single mobile resource she can move along an arbitrary graph to protect some valuable vertices, called \emph{targets}. Moreover, each target is covered by an alarm system capable of raising a different alarm signal for every target that is under attack. 
On the other side, there is an \textit{Attacker}, equipped with multiple attacking resources and able to observe the actions undertaken by the Defender. 
The goal of the Attacker is to perform the highest damage, employing the resources to perform any type of attack, adopting any sequential strategy or using all of them simultaneously. This latter case exactly reflects what happened during the terroristic attacks in Paris~2015, as argued below.

\subsection{Motivating Scenario}
\begin{figure}[htbp]
	\centering
	\subfigure[Paris map]{\includegraphics[scale=0.35]{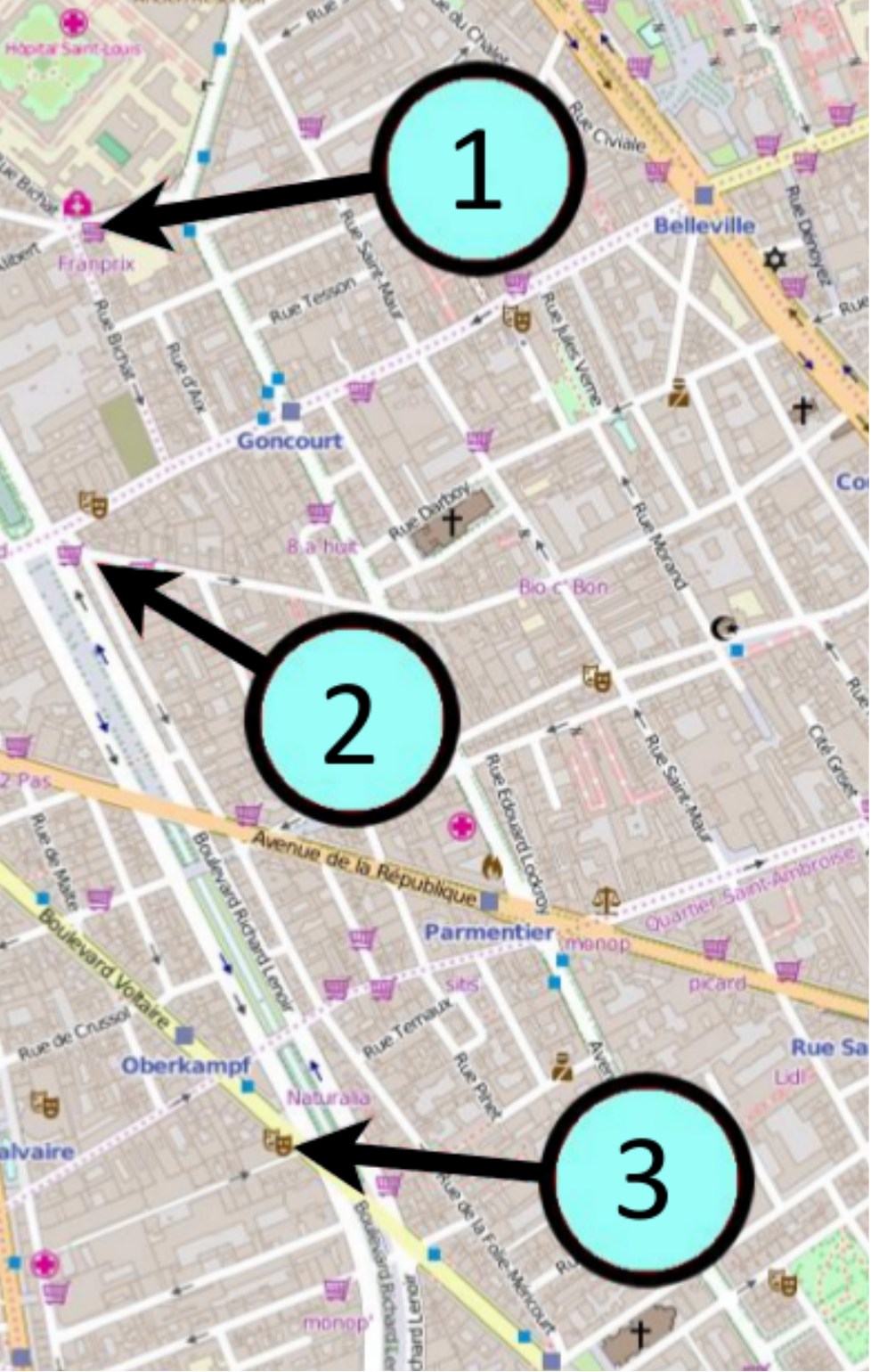} \label{fig:paris_map}}
	\qquad \quad
	\subfigure[Paris graph]{\includegraphics[scale=0.5]{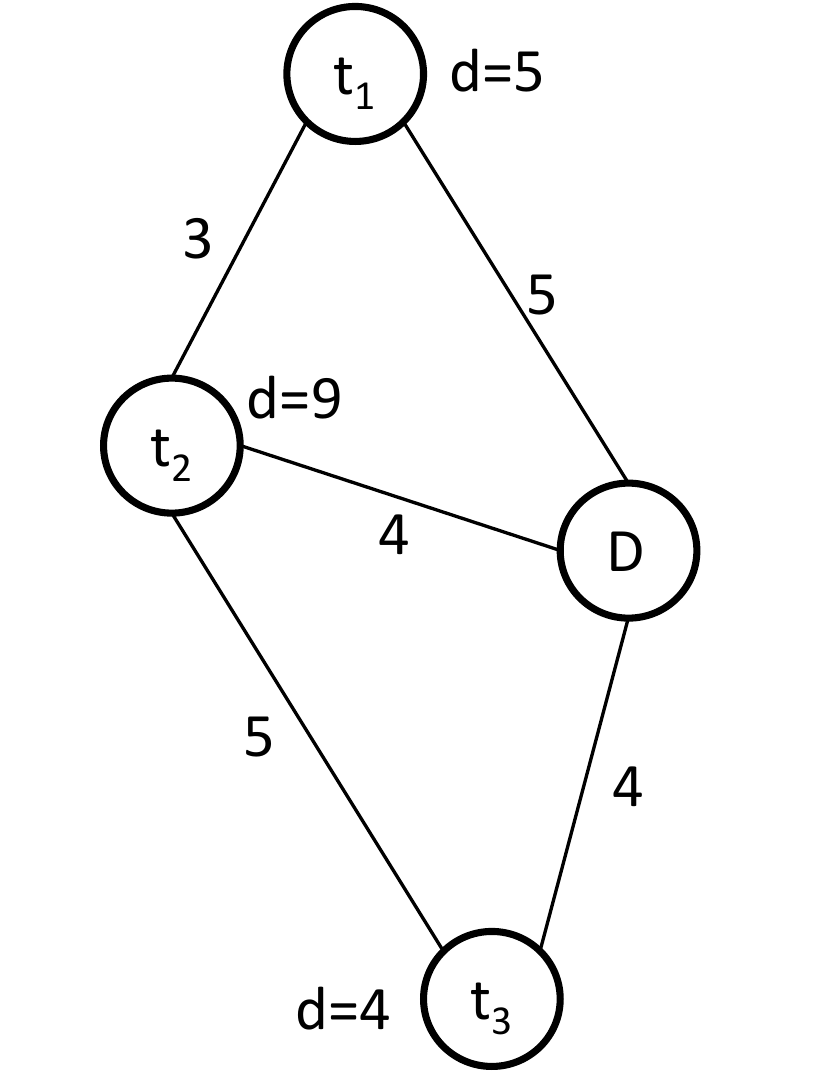} \label{fig:paris_graph}}
	\caption[Terroristic attacks in Paris, 2015]{Terroristic attacks in Paris, 2015.}
	\label{fig:paris}
\end{figure}

To illustrate how our model can be used to tackle real-world problems, we propose a simple but concrete example that perfectly fits, namely the series of terroristic assaults that affected Paris in 2015. On November 13, 2015, in Paris, six attacks were performed across all the city: one near the Stade de France, three very close to each other, at the intersection of Rue Bichat and Rue Alibert, in Rue de la Fontaine-au-Roi and at the Bataclan theatre, and the last two in Rue de Charonne and in Boulevard Voltaire. For simplicity, we focus on the three that happened close to each other, both in space and time. 
Figure~\ref{fig:paris} shows the locations in which these three attacks happened: specifically, Figure~\ref{fig:paris_map} shows the actual map of the city of Paris, while Figure~\ref{fig:paris_graph} shows a possible graph representing the part of the city on which our game could be played. Notice that the longest distance, namely the one between \textcircled{$1$}, at the intersection between Rue Bichat and Rue Alibert, and \textcircled{$3$}, The Bataclan theatre, is about 1150 meters. 

To model the situation represented in Figure~\ref{fig:paris}, we designed in Figure~\ref{fig:paris_graph}, $t_1, t_2, t_3$ as the attacked locations (targets), while $D$ is the starting point of the Defender. The weights of the edges provide the distances among the various places while the penetration times $d$, i.e., the time needed by the Attacker to complete the attack, have been set taking into account how easy it could be to hide and run away after having committed a crime in those places.

\subsection{Original Contributions}
In this work, we tackle the problem of facing multiple attacks in adversarial patrolling games when the graph is arbitrary and utilities are zero-sum.
Given such setting, our goal is to provide the Defender with the optimal strategy to protect the valuable nodes.
Although the model we propose can be refined along a number of dimensions, e.g., uncertainty over the alarm signals (introducing false negatives/positives) or multiple defensive resources, we show that even the study of the basic version is involved and completely unexplored in the literature so far. To the best of our knowledge, sequential attacks, intended as different attacks happening in the same game, are studied only in patrolling games with very restricted environments~\cite{sless2014multi} and in cyber-security games, where no graph constraints are present~\cite{zhao2016optimizing}. 
The challenge of our task is due to the high interaction level among the players, e.g., the Attacker could use some resources to make the patroller  move away from some valuable targets and, subsequently, carry an attack against those targets once the Defender has run away.
In principle, an equilibrium can be found in polynomial time in the size of the game tree, but, here, the game tree induced by the model is exponentially large in the size of the graph and the number of resources available to the Attacker. 
When the number of resources is a fixed parameter, the problem admits an algorithm capable of finding the strategies \emph{on the equilibrium path} requiring polynomial time in the size of the graph.
Conversely, we show that there is no algorithm requiring polynomial time in the number of Attacker's resources, unless $\mathsf{P}=\mathsf{NP}$, even in the simplified case in which the Attacker uses all her resources simultaneously. 
The computation of the equilibrium strategies requires the common knowledge about the number of Attacker's resources
However, this information is unlikely to be common. Customarily, in that case, Bayesian games are used, but this approach appears unsuitable here where it would require that the game tree contains a subtree for each possible number of resources of the Attacker. For this reason, we investigate a different problem: the robustness of a Defender's strategy when the guess about the number of resources the Attacker can employ is wrong. We evaluate the worst-case inefficiency of this strategy showing that can be arbitrary even when the guess is a wrong estimate---both over and under---for just a single resource. This happens when the inefficiency is defined as the ratio of the Defender's utilities obtained with a wrong guess and a correct guess. When instead the inefficiency is defined as the difference between those utilities, in an $\epsilon$-Nash equilibrium fashion, the higher the error in the estimation, the higher the loss.
Furthermore, we investigate the use of \emph{online} algorithms to adopt when no information is available to the Defender. We provide a tight upper bound over the \emph{competitive factor} when non-stochastic online algorithms are used, and we show that the competitive factor can be improved by resorting to randomization.

\subsection{Structure of the Paper}
The rest of the paper is organized as follows. 
Section~\ref{sec:related_works} presents a brief state of the art of Security Games and discusses the principal works related to ours, while Section~\ref{sec:background} introduces the basic model on which we build ours and the results we exploit from the literature.
Section~\ref{sec:problem_formulation} presents our model, characterizing its main features and introducing the presence of multiple attacking resources. 
Then, we focus on the Attacker, who may perform multiple attacks: Section~\ref{sec:simultaneous_attacks} studies what happen if she can perform simultaneous attacks while Section~\ref{sec:sequential_attacks} generalizes such cases, enabling the Attacker to perform different sequential attacks. 
In Section~\ref{sec:no_prior}, we investigate scenarios in which the Defender does not know \textit{a priori} the number of resources available to the Attacker.
Finally, Section~\ref{sec:conclusions} concludes the work and presents the future research lines that we will investigate.
Section~\ref{appendix:notation} provides a table summarizing the main symbols adopted throughout the paper.

\section{State of the Art} \label{sec:related_works}
Here we introduce the principal works related to ours.
First, we briefly introduce \emph{Security Games} in Section~\ref{sec:security_games}, then in Section~\ref{sec:patrolling_games} we focus on Patrolling Games, ours being one of them. Finally, in Section~\ref{sec:most_related}, we present the works that are the most related to ours, except the works~\cite{basilico2017coordinating,basilico2016security,basilico2017adversarial}, which we discuss in detail in Section~\ref{sec:background} where we introduce our patrolling model.

\subsection{Security Games} \label{sec:security_games}
The birth of the first \emph{Security Game} (SG) is due to John Von Neumann, with the \emph{Hide and Seek Game}~\cite{flood1972hide}. The scenario is the following: a player hides in a place among a finite number of them, and the (unique) opponent should find her. It is modeled as a normal-form zero-sum game, as the goals of the two players are precisely one the opposite of the other.

From this simple game, many directions have been investigated, originating a lot of papers in the following years, where some fugitives are escaping from pursues that want to reach them~\cite{adler2003pursue}. If the fugitive tries to reach a target, e.g., a vanishing point, while the pursuer has to stop her, we have \emph{Ambush Games}~\cite{ruckle1976ambush} while if the fugitive hides and the pursuers look for her, we have \emph{Search Games}~\cite{gal1980search}. Finally, if both fugitives and pursues can \emph{move} in the environment, we call them \emph{Infiltration Games}~\cite{alpern1992inf}.

These studies evolved in research about strategic resource allocation for security, which has been a very prolific domain in the field of algorithmic game theory during the last years. The investigation in this domain led to the development of what today are commonly called {\em Security Games (SGs)}: game-theoretical frameworks for computing resource allocation strategies against adversarial security threats. Security is one of the most critical issues every country and every person deals with every day: the protection of airports, ports, banks, monuments, and museums, but also containing urban attacks, controlling poaching of endangering species, preventing the diffusion of misinformation and guaranteeing cybersecurity~\cite{kar2017trends}.

Customarily, SGs are a mathematical tool to model the protection of infrastructures or open environments as a non-cooperative game between a \emph{Defender} and an \emph{Attacker}. Given the setting, these scenarios take place under a \emph{Stackelberg} (a.k.a. \emph{leader-follower}) paradigm~\cite{von2004leadership}, where the Defender (\emph{leader}) commits to a strategy and the Attacker (\emph{follower}) first observes such commitment, then best responds to it.
From a computational perspective, as discussed in~\cite{conitzer2006computing}, finding a leader-follower equilibrium is computationally tractable in games with one follower and complete information, while it becomes hard in Bayesian games with different types of Attacker. The availability of such computationally tractable aspects of Security Games led to the development of algorithms capable of scaling up to huge problems, making them deployable in the security enforcing systems of several real-world applications. 

There have been several applications based on such games, and we describe some of them. The first one to be deployed is ARMOR, consisting in the strategic placement of checkpoints on the streets leading to the Los Angeles International Airport (LAX) and the management of the patrolling units across the terminals~\cite{paruchuri2008playing,pita2008deployed}. The authors cast the problem as a Bayesian Stackelberg game, giving the guards the possibility to assign different and appropriate weights to their actions and tune them with respect to  the types of the adversary.
In~\cite{tsai2009iris}, the problem of scheduling undercover air marshals on U.S. domestic flights has been tackled with the project IRIS. Here, additional constraints have been introduced, since the agents must fly among cities such that, the next day, they will depart from the same city they landed the day before. Moreover, the agents are scheduled to have a list of cities such that the first and last cities are the same so that they actually fly around following a circle.

In the same year, new models and algorithms were designed to deal with more complex and more realistic instances~\cite{kiekintveld2009computing}. The authors proposed a compact way of representing a security game, thus obtaining an exponential improvement both in the running time and the memory space needed by the algorithms to solve such problems.

In 2012, game-theoretic techniques were adopted to secure ports, bridges, and ferries, giving the U.S. Coast Guard patrols the best tours they should follow to protect such areas~\cite{an2012protect}. Since patrols have a fixed duration during the day, it was not possible to model each target as a node in a graph, as it is customarily done, because of the size of the problem, having a large number of targets to be considered. The solution was to define patrol areas, i.e., groups of nearby targets, which lead the Coast Guard to redefine the set of defensive activities to be performed. Moreover, this is one of the first deployed projects in which there is an assumption of non-perfect rationality of the Attacker.

GUARDS is another prominent project, which builds upon ARMOR and IRIS~\cite{pita2011guards}. The goal is the protection of an airport at a national scale. The authors introduced the possibility of dealing with hundreds of heterogeneous security activities to be able to prevent several potential threats with a system designed for hundreds of end-users.

With TRUSTS~\cite{delle2014game}, the authors designed a security game to schedule city guards to stop fare evasion in Los Angeles Metro. Each day, TRUSTS generates a patrol schedule for a team of inspectors, consisting of a sequence of fare-check operations, alternating between in-station and on-train operations. Each operation indicates specifically where and when a patrol unit should check fares. They used  Markov Decision Processes (MDPs) as a compact representation to model each Defender unit's patrol actions.

In addition to these projects, which are still deployed, Security Games have been adopted for many different situations. With respect to  airport security, new games have been studied to perform an effective screening for threats, checking both objects and people before they enter the airport (similarly, they could be applied to a cargo container or stadium screening)~\cite{brown2016one,schlenker2016get}. The challenge is to find a dynamic approach for randomized screening, allowing for more effective use of limited resources while improving the level of security. The authors designed Threat Screening Games (TSGs), where there is a set of scarce resources to screen and check several individuals or objects. The proposed approach, GATE, applies to Bayesian general-sum TSGs.

The protection of water from toxic materials has been studied in~\cite{ford2016nectar}. In some countries, despite government regulations on leather tannery waste emissions, inspection agencies do not have enough resources to control the problem, and tanneries' toxic wastewaters have a destructive impact on surrounding ecosystems and communities. NECTAR has been proposed as the first security game application to generate environmental compliance inspection schedules. Still related to water-life, in~\cite{yin2016efficient}, the authors proposed a method to protect coral reefs, which are valuable and fragile ecosystems, constantly under threat from human activities, e.g., coral mining. Many countries have built marine protected areas and protect their ecosystems through boat patrol, and efficiently schedule these patrols is a perfect application for security games.

In~\cite{gholami2016divide}, a new game to study the behavior of the Defender with respect to  multiple independent adversaries is proposed. In fact, collusion among malicious agents is very common in every domain, including airports and wildlife security. The authors study whether it could be more convenient for a group of Attackers to break up collusion by playing off the self-interest of individual adversaries.

Preventing crimes or terrorist attacks in urban areas is the problem that has been tackled in~\cite{zhang2017optimal}. Guards must respond very quickly to be able to intercept and catch a potential Attacker on her escaping route, which could depend on time-dependent traffic conditions on transportation networks. The primary challenge here consists of the presence of time constraints both on the Defender and the Attacker side.

Very recently, Security Games have been applied to stop the nuclear smuggling in international container shipping through advanced inspection facilities~\cite{wang2017stop}. Efficiency and efficacy are fundamental for this task, given that there are millions of containers, which should be screened. This work models the interaction between an inspector and a smuggler using of a security game, formulating the smuggler's sequential decision behavior as a Markov Decision Process.

All the above works necessarily work with a discrete and finite number of strategies per player, often not explicitly taking into account the underlying topology of the space in which targets are located. However, in many practical security settings, defense resources can be located on a continuous plane, and so defense solutions are improved by placing resources in a space outside of actual targets (e.g., between targets). To address this limitation, the authors proposed Security Game on a Plane, where targets and defensive resources are distributed on a 2-dimensional plane, able to protect targets within a certain effective distance~\cite{gan2017security}.

The papers listed above deal with physical security. Recently, game-theoretic techniques have also been applied to \emph{cyber} security. In~\cite{durkota2015approximate,durkota2015optimal}, the authors study the problem of protecting a network in which an administrator may decide the best security measures to use to improve the safety of the network. This is achieved by resorting to honeypots, i.e., decoy services or hosts, placed by the Defender, while the Attacker chooses the best response as a contingency attack policy.

Still in a cybersecurity dimension,~\cite{schlenker2017don} proposes an approach to investigate whether alerts generated by potential cyber attacks are real attacks or just false positives. Also here, the magnitude of the problem is high, with a number of alerts that is overwhelming with respect to  the number of analysts that can check the authenticity of such attacks.

It is immediate to observe that Security Games have tons of applications. In the next section, we focus on Patrolling Games, i.e., games in which the Defender controls a mobile resource that can catch the Attacker.

\subsection{Patrolling Security Games} \label{sec:patrolling_games}
In this work, we focus on a specific class of security games, called \emph{Patrolling Security Games} (PSGs). These games are modeled as infinite-horizon extensive form games in which the Defender controls one or more {\em patrollers} moving within an environment, represented as a finite graph. In~\cite{basilico2012patrolling}, the authors provide the formulation of a Security Game in which the Defender controls a mobile resource moving in the environment between adjacent areas, while the Attacker can observe the movements of the patrollers at any time and use such information in deciding the most convenient time and target location to attack

When multiple mobile resources are available to the Defender, coordinating them at best is, in general, a hard task which, besides computational aspects, must also keep into account communication issues~\cite{basilico2010asynchronous}. In this work, the authors determined the smallest number of robots needed to patrol a given environment and computed the optimal patrolling strategies along several coordination dimensions, e.g., the strategy of a robot can or cannot depend on the strategies of the other robots and the environment can or cannot be partitioned.

However, the patrolling problem is tractable, even with multiple patrollers, in border security (e.g., linear and cycle graphs), when patrollers have homogeneous moving and sensing capabilities and all the vertices composing the border share the same features~\cite{agmon2011multi}. Scaling this model involved the study of how to compute patrolling strategies in scenarios where the Attacker is allowed to perform multiple attacks~\cite{sless2014multi}. Similarly, coordination strategies among multiple Defenders are investigated in~\cite{agmon2012coordination}. 

In~\cite{an2014adp}, the authors study the case in which there is a temporal discount on the targets, i.e., the value of the targets diminishes as time passes by, both for the Attacker and the Defender. 
Extensions are discussed in~\cite{shieh2013efficiently}, where coordination strategies between Defenders that must execute joint activities are explored, in~\cite{gan2015security}, where a resource can cover multiple targets, and in~\cite{agmon2010events} where attacks can be detected at different stages with different associated utilities. 

Patrolling has always been a fundamental research line in robotics: in~\cite{amigoni2009finding}, for the first time, a security game has been explicitly cast in this domain. Protecting sites against intrusions is a topic of increasing importance, and robotic systems for autonomous patrolling have been developed in the last years. However, unpredictable strategies are not always efficient in getting the patroller a large expected utility. Here, exploiting a model of the adversary in a game theoretic framework, the authors provide a method to find the optimal strategies, modeling a given patrolling situation as an extensive-form game. In~\cite{basilico2009capturing}, the model is refined, capturing patroller's augmented sensing capabilities and a possible delay in the intrusion. In~\cite{amigoni2010moving}, the authors conducted realistic experiments by using USARSim~\cite{carpin2007usarsim}, to study the behavior of the optimal patrolling strategy both in situations that violate its idealized assumptions and in comparison with other patrolling strategies.

A significant contribution comes from~\cite{basilico2011automated}, where it is proposed the first study on the use of abstractions in security games (specifically for PSGs) to design scalable algorithms. The authors defined some classes of abstractions and provided parametric algorithms to automatically generate such abstractions, which allow one to relax the constraint of patrolling strategies' Markovianity, a usual assumption in PSGs, and to solve large game instances.

Higher degrees of interaction between the players by means of a sequential structure were explored, e.g., in~\cite{alpern2011patrolling,papadaki2016patrolling}: the authors analytically determined the value of the game or bounds on the value, for various classes of graphs, especially where the network is a line, which models the problem of guarding a channel or protecting a border from infiltration.

Finally,~\cite{munoz2013introducing} proposes a first skeleton model of an alarm system where sensors and the authors analyze how sensory information can improve the effectiveness of patrolling strategies in adversarial settings with the Attacker able to perform a single attack. They show that, when sensors are not affected by false negatives and false positives, the best strategy prescribes that the patroller responds to an alarm signal rushing to the target under attack without patrolling the environment. As a consequence, in such cases the model treatment becomes trivial.

\subsection{Related Works} \label{sec:most_related}
Here, we present the works that are closely related to ours. 

\paragraph{Green Security Games (GSGs).} These games constitute a novel game model where a generalized Stackelberg assumption is made~\cite{fang2015security}. As it happens in our game model, the Attacker can perform multiple attacks. However, GSGs are repeated games in which, at each repetition, the same game is played. Differently, in our game model, players play a unique (non-repeated game) on a game tree. Furthermore, we adopt the common full-rationality assumption made in game-theoretic frameworks, while in GSGs a bounded rationality assumption is made. Such assumption is central in the application domain to which Green Security Games apply. Conversely, our primary focus is on studying the worst case, thus playing against a rational adversary. Our work is different from works dealing with such problem, e.g.,~\cite{nguyen2016capture}: in our model, as it is common in real-world applications, the Attacker needs multiple turns to conquer a target while in~\cite{nguyen2016capture} only one-shot attacks are considered, without the possibility of deceiving the Defender.

\paragraph{Stochastic Games.} Our game model would reduce to a stochastic game~\cite{chatterjee2012survey} if we force that the strategies of the players to depend on a history of observations that is somehow bounded, e.g., depending only on the current vertex in which the patroller is. However, in this case, the optimal strategy---that can be obtained by solving a stochastic game---could be arbitrarily inefficient with respect to  the optimal unconstrained strategy.

\paragraph{Attack Graph Games (AGGs).} This class of games exploits a particular structure, called \textit{attack graph} (AG), to represent a vast space of sequential Attacker's actions. Specifically, an AG is a directed AND/OR graph consisting of fact nodes F (OR) and action nodes A (AND), where every action node has preconditions, i.e., facts that must be true before the action can be performed, and effects, namely a set of facts that become true if the action is successfully executed. These relations are represented by edges in the attack graph~\cite{durkota2015approximate,durkota2015optimal}. However, as we have already pointed out, the graph in our model represents the environment in which the game is played and not just the possible actions of the Attacker. Moreover, the authors consider the detection of cybersecurity attacks using honeypots, which are static; conversely, our defending resource can patrol among the areas of the environment.

Notice that none of the previous works considers the introduction of an alarm system capable of providing additional information to the Defender. Thanks to an alarm system, for the first time, the patroller can exploit dynamic information against multiple attacks that can be carried out sequentially, moving according to how the attacks are performed. Conversely, all the other works rely on observations made during the patrolling and other prior knowledge.

\section{Background} \label{sec:background}
The problem we study builds upon some results provided in~\cite{basilico2017coordinating,basilico2016security,basilico2017adversarial}. We briefly introduce the basic model in Section~\ref{sec:basic_model}, while in Section~\ref{sec:previous_results} we report the main results available in the literature that we exploit in our paper.
Section~\ref{sec:challenge} discusses how the results presented in our work relate to the results known in the literature.

\subsection{Basic Model} \label{sec:basic_model}
There is an environment to be patrolled, modeled as a graph, in which the vertices represent different areas of the environment and the edges represent the connections among such areas. All the edges require one turn to be traversed. We define the set of targets as the set of valuable nodes, characterized by a value and a penetration time, i.e., the time needed to be compromised. An \emph{alarm system} generates a signal whenever a target is under attack.

A 2-player security game is played by an Attacker~\Att~and a Defender \Def. In this game, \Att~seeks to gain value by compromising some targets while \Def~controls one single patroller by specifying a movement strategy for it. The game can be formulated as an extensive-form infinite-horizon zero-sum game, with \Att~and \Def~playing alternatively. Each turn is constituted by one action for the Attacker and the subsequent action for the Defender.

\subsection{Previous Results} \label{sec:previous_results}
In~\cite{basilico2017adversarial}, the alarm system is affected by \emph{spatial uncertainty}, i.e., the alarm system is uncertain about the exact target under attack. The actions of the Attacker correspond to the targets, while the actions of the Defender are the so-called covering routes, i.e., finite sequences of vertices such that each target traversed while following the routes is reached within its penetration time.

Given that there are no false positives nor false negatives, the problem can be split into two games, namely the \textit{Signal Response Game} (SRG) and \textit{Patrolling Game} (PG). The SRG captures the situation in which the Defender is in some vertex $v$ and the Attacker attacked a target, while the PG models the case in which the Defender moves in the absence of an alarm signal. 
The authors proved that solving an SRG is \textsf{NP}-hard even with a single signal.
In the PG, the best strategy results standing in a vertex, waiting for an alarm signal, and best responding to it. Finding a patrolling strategy for the Defender is \textsf{FNP}-hard if the graph is a tree and \textsf{APX}-hard if the structure of the environment is arbitrary. 
The authors proposed an exact algorithm (\texttt{SolveSRG}) whose complexity is $O(2^{|T|}\cdot|T|^5)$, where $|T|$ is the number of targets in the graph. 

In~\cite{basilico2017coordinating}, the authors study the scenario in which the Defender is allowed to control multiple resources and alarms are affected only by spatial uncertainty. As in the case of a single resource, the best strategy of the Defender is to strategically place the resources, wait for a signal and then move them accordingly. The authors study the computational complexity of finding the minimum number of resources needed to protect an environment, i.e., the resources are located such that no target is far from a resource more than its penetration time. Furthermore, the authors provide exact and approximation algorithms to find the best Defender's strategy.

In~\cite{basilico2016security}, the authors focus on the case in which the Defender has a single resource, and the alarm system is affected by false negatives, such that, even though an attack is carried on, an alarm signal may not be raised. In this scenario, standing in a vertex until an alarm arises may be an arbitrarily inefficient strategy. Instead, the best strategy may prescribe that the patroller also moves before some signal arises. The authors focus on the study of strategies that can be computed in practice. 

\subsection{New Challenge} \label{sec:challenge}
In Table~\ref{table:results}, we classify the results already known in the literature together with our original contributions, denoted with `$\times$'.
	
\begin{table}[!h]
	\begin{center}
		\hspace{-0.3cm}\begin{tabular}{|c|c|c|c|}
			\hline
			& Perfect alarm & Spatial uncertainty & Spatial uncertainty \\ 
			& 			 & 				 &  and false negatives\\ \hline
			Single \Def--Single \Att & \cite{basilico2017adversarial} & \cite{basilico2017adversarial} & \cite{basilico2016security} \\ \hline
			Multi \Def--Single \Att & \cite{basilico2017coordinating} & \cite{basilico2017coordinating} &  \\ \hline
			Single \Def--Multi \Att & $\times$ &  &  \\
			\hline
		\end{tabular}
		\caption{Known results and our contribution (denoted by `$\times$').}
		\label{table:results}
		\vspace{-0.8cm}
	\end{center}
\end{table}

\section{Problem Formulation} \label{sec:problem_formulation}
In this section we introduce our model: Section~\ref{sec:game_description} introduces the patrolling setting, while Section~\ref{sec:game:interaction} describe the game mechanism.

\subsection{Patrolling Setting} \label{sec:game_description}
Our game is modeled by the model introduced in Section~\ref{sec:basic_model}. Differently, from that model, we allow the Defender \Def~to exploit information gained from a \emph{perfect} alarm system and the Attacker \Att~to control $k$ resources, which can be employed simultaneously or sequentially. We model the environment as a graph $G=(V,E)$ with unitary edges. $\omega^*_{i,j}$ denotes the smallest traveling cost in turns between vertices $i$ and $j$. $T \subseteq V$  is the set of targets characterized by a value $\pi(t)\in (0,1]$ and a penetration time $d(t)\in \mathbb{N}^+$. A perfect alarm system  generates a signal $s_i$ \emph{if} and \emph{only if} target $t_i$ is under attack. Any generated signal $s_i$ is common knowledge. Since each signal corresponds exactly to one target and \emph{vice versa}, we can safely refer to the signals triggered by the alarm system directly by the targets under attack. The Attacker can use $k$ resources, while \Def~controls one single patroller. 

\subsection{Game Mechanism} \label{sec:game:interaction}
Attacker \Att~and Defender \Def~play alternatively in an extensive-form infinite-horizon zero-sum game. Each turn is constituted by one action for the Attacker and the subsequent action for the Defender.
At each turn $\tau$, the Attacker may decide to \emph{wait} or to \emph{attack}, with an attack being characterized by the pair $(\tau, \mathsf{attacked}(\tau))$ where $\tau$ is the turn at which the attack begins\footnote{We assume \Att~can instantly reach the attacked target. This can be relaxed as shown in~\cite{basilico2009capturing}.} and $\mathsf{attacked}(\tau) \subseteq T$ is the \emph{the support of the attack}, i.e., the set of the attacked targets. Once \Att~has employed a resource to make an attack, such a resource lays on the target until the attack is concluded. Moreover, each resource can be employed just once by \Att. On the other hand, \Def~observes the signals triggered by the alarm system (if any) and decides whether to keep its resource in the same area or to move it along the graph. We assume the Defender places her patrolling resource in the environment before the first attack is performed. The choice of such placement is part of the solution to the problem. The utilities of \Def~and \Att~are $\left(-\sum_{i=1}^k \gamma_i \, \pi(t_i), \sum_{i=1}^k \gamma_i \,\pi(t_i)\right)$, where $\gamma_i = 0$ if \Att~attacks target $t_i$ at $\tau$ and the patroller traverses $t_i$ by $d(t)$ turns after~$\tau$, catching the attacking resource, otherwise, if \Att~completes the attack on target $t_i$ without being detected, $\gamma_i = 1$. Once a resource of \Att~attacking target~$t$ has been detected by \Def, the resource is discarded from the game and, in principle, \Att~can attack target~$t$ again in future using another resource (if any available). Similarly, a target can be successfully compromised only once, after that it is considered as a vertex without any value. Finally, if \Def~protects the environment from all the attacks, her utility is zero, corresponding to the maximum utility she can get. Otherwise, for each target successfully compromised, \Def~loses the value of such target.

In the following sections, first, we study the restricted case in which \Att~deploys all the resources simultaneously. Notice that this restriction induces the game to be finite and thus the equilibrium can be computed before the game is played. Subsequently, we study the unrestricted case. Remarkably, in that case, the game tree may be arbitrarily large---\Att~may wait indefinitely before making an attack---and therefore there is no way to find an equilibrium before the execution of the game. Nevertheless, we show that, once $k$ is fixed, there is a polynomial time algorithm to find the equilibrium path. Thus, for small values of~$k$, \Def~can compute the equilibrium path before the play and apply it and, if \Att~behaves irrationally not following the equilibrium path, \Def~can compute on-the-flight the equilibrium path of the subgame she is playing.

\section{Facing Simultaneous Attacks} \label{sec:simultaneous_attacks}
First, we study the restricted case in which \Att~attacks employing all the $k$~resources simultaneously. Tackling such problem is functional to solve the general case with sequential attacks. Since \Att~does not pay any cost to use the resources, it easily follows that she will use all of them, each for a different target. When the attacks take place, being simultaneous, $k$ signals will be raised, and the Defender must compute a path along the graph to protect the corresponding targets. W.l.o.g., we assume the attack to begin at~$\tau=0$. In this case, we can safely adopt covering routes as actions for the Defender\footnote{This holds because, as it will be proved in Theorem~\ref{thm:sa_hard}, the problem of protecting targets from a single attack with a spatial uncertain alarm system can be mapped to the problem with a punctual alarm system and simultaneous attacks.}. We introduce the formal definition.

\begin{definition}[\textit{Direct route}]
	Given $\mathsf{attacked}(0)$, a {\em direct route} is a sequence $r=(r(0), r(1),$ $ \ldots, r(h))$ of arbitrary finite length $h$, where $r(0)$ is any vertex of $G$ and $r(i)$ is any target in $\mathsf{attacked}(0)$.
\end{definition}

A direct route can be instantiated to a graph walk starting from $r(0)$ and traveling any shortest path between $r(i)$ and $r(i + 1)$. For any $i \in \{1, \ldots, h\}$, call $A(r(i)) = \sum_{l=0}^{i-1}\omega^*_{r(l),r(l+1)}$ the time required by the walk to go from $r(0)$ to $r(i)$. Notice that here we call \textit{direct routes} sequences of nodes that in Section~\ref{sec:previous_results} were called \textit{routes}. This is needed since we will have to generalize the concept of route in the following (see Section~\ref{sec:two_seq_attacks}, Definition~\ref{def:route_general}).

\begin{definition}[Covering route] 
	A direct route $r$ is a {\em covering route}, denoted as $r^c$, if $\,\forall i \in \{1, \ldots, h\}$, it holds $A(r(i)) \leq d(r(i))$.
\end{definition}

Any other target $t$ not appearing in the direct route is not visited or visited after $d(t)$ turns from the start of the attack.

The resolution approach for $k=1$ is easy, being a sub-case of the problem studied in~\cite{basilico2017adversarial}. More precisely, the best strategy for the patroller is to stay on a vertex~$v$, wait for a signal~$s$ associated with a target~$t$ and, when raised, move towards~$t$ along the shortest path connecting $v$ to $t$. The problem can be solved in polynomial time in~$|V|$.

To deal with multiple attacking resources, we first have to figure out the space of the actions available to \Att~and \Def. The Attacker can attack any subset of $k$ targets, i.e., her actions are all the possible combinations of $k$ targets among $|T|$ targets, $\binom{|T|}{k} \approx |T|^k$ in total. On the other side, the Defender must compute the covering routes for the $\mathsf{attacked(0)}$ targets under attack. As it can be seen, the space of the actions is exponential both on the side of \Def~and of \Att. Thus, it is natural to wonder whether we have to enumerate all of them or if we can find a compact way to express them. On the Attacker side, we can adopt marginal strategies, i.e., \Att~plays directly on the targets, and then, exploiting the Birkhoff-von Neumann theorem~\cite{birkhoff1946tres}, as done in~\cite{korzhyk2010complexity}, we map the correlated strategy back to a feasible mixed strategy. Thus, her space of actions can be exponentially compressed. Conversely, nothing can be done for the Defender, being the computation of a covering route a \textsf{NP}-hard problem also in our novel setting and therefore there is no algorithm running in polynomial time in~$k$, unless $\mathsf{P}=\mathsf{NP}$.

\begin{definition}[Simultaneous-Attack problem (SA-$v$)] 
	The Simultaneous-Attack problem is defined as follows.
	
	\begin{itemize}
		\item INSTANCE: an instance of our problem with the patroller in a given vertex $v$, with $\mathsf{attacked}(0)$ targets attacked simultaneously by \Att;
		\item QUESTION: does $\mathsf{attacked}(0)$ admit any covering route $r^c$?
	\end{itemize}
\end{definition}

\begin{theorem} \label{thm:sa_hard}
	SA-$v$ is \textsf{NP}-hard.
\end{theorem}

\noindent
\textit{Proof.} We provide a reduction from COV-SET, which is proved being $\mathsf{NP}$-hard in~\cite{basilico2017adversarial}.

\begin{definition}
	The COV-SET problem is defined as:
	\begin{itemize}
		\item INSTANCE: an instance of SRG-$v$ with a target set $T$;
		\item QUESTION: is $T$ a covering set? (Equivalently, does $T$ admit any covering route $r$?)
	\end{itemize}
\end{definition}

\textit{Mapping}. We map an instance of COV-SET to an instance of SA-$v$ by constructing $\mathsf{attacked}(0) = T' = \{t_1, t_2, \ldots, t_h\}$ and associating to each $t_i \in \mathsf{attacked}(0)$ a unique signal $s_i$.

\textit{If}. If SA-$v$ admits a covering route for $\mathsf{attacked}(0)$, this means all the targets in $\mathsf{attacked}(0)$ can be covered within their penetration times and thus, by construction, also $T'$ admits a covering route. 

\textit{Only if}. It can be proved following steps similar to the \textit{If} direction. If $T'$ admits a covering route, then there is also a feasible covering route for $\mathsf{attacked}(0)$.
\hfill $\Box$

Finally, we observe that, when the starting vertex $v$ and $\mathsf{attacked}(0)$ are fixed, the problem can be solved in $O(\,2^k\, k^5\,)$, with $k$ being the size of $\mathsf{attacked}(0)$, while in general the problem has a higher complexity, equal to $O(\,|V|\,2^k\, k^5\,\binom{|T|}{k}\,)$, which approximately is $O(\,|V|^{k+1}\,2^k\, k^5\,)$. This follows from the algorithm \texttt{DP-ComputeCovSets}, called \texttt{SolveSRG}($v,T'$) from now on, proposed in~\cite{basilico2017adversarial}, where $v$ is the starting vertex and $T'$ the set of attacked targets, whose complexity in general is $O(2^{|T'|}\, |T'|^5\,)$. Furthermore, the following result holds.

\begin{theorem}
	SA-$v$ problem is $\mathsf{NP}$-hard on tree graphs.
\end{theorem}

The proof follows from the reduction reported in the proof of \cite[Theorem 1]{basilico2017adversarial}, which exploits instances with a single signal, and can be thus directly applied to our case.

\section{Facing Sequential Attacks}\label{sec:sequential_attacks}
First, we ask whether it is worth studying the case in which the Attacker can perform sequential attacks. To answer this question, we show that \Att~can gain strictly more from sequential attacks with respect to  simultaneous attacks, as stated in the next proposition.

\begin{proposition} \label{prop:multi_att}
	There exist patrolling games that can provide a strictly higher utility to the Attacker when she performs multiple sequential attacks to the targets rather than a single attack to multiple targets.
\end{proposition}

\noindent
\textit{Proof.} The proof is given by example. Consider the following graph, where $d(t_1)=d(t_2)=4$, edges are unitary and $k=2$.

\begin{figure}[!htbp]
	\centering
	\begin{tabular}{c}
		\begin{tikzpicture}
		[scale=.8,auto=left,every node/.style={circle,fill=white,draw,minimum size = 0.675cm,font=\sffamily\large\bfseries}]
		\node (n1) at (0,0) {};
		\node (n2) at (-4,0) {$t_1$};
		\node (n3) at (-2,0) {};
		\node (n4) at (2,0) {};
		\node (n5) at (4,0) {$t_2$};
		
		\path[every node/.style={font=\sffamily\small}]
		(n1) edge node [loop] {} (n3)
		(n1) edge node [loop] {} (n4)
		(n2) edge node [loop] {} (n3)
		(n4) edge node [loop] {} (n5);	
		\end{tikzpicture}
	\end{tabular}
	\caption{Linear graph employed in the proof of Proposition~\ref{prop:multi_att}.}
	\label{fig:seq_needed}
\end{figure}
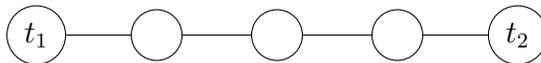

If \Att~can perform only simultaneous attacks to the two targets, \Def~will stay in $t_1$ (or $t_2$), able to reach $t_2$ (or $t_1$) within its penetration time. Hence, \Def~will protect both targets.  Conversely, if \Att~can attack the two targets sequentially, it can be observed that, no matter her position, the Defender can only protect one of the targets, losing the other. For instance, if \Def~stays in $t_1$ (or $t_2$), \Att~will attack $t_2$ (or $t_1$). Subsequently, if \Def~does not move to cover $t_2$ (or $t_1$), then the target under attack is lost. Otherwise, if \Def~moves to cover $t_2$ (or $t_1$), then \Att~will attack $t_1$ (or $t_2$) immediately after the first move of \Def~and thus \Def~cannot cover both targets. 
\hfill $\Box$

In the rest of the section, we investigate the best defense strategy for the Defender when the Attacker can perform sequential attacks.
Specifically, Section~\ref{sec:two_seq_attacks} tackles the problem when the Attacker can perform two sequential attacks, i.e., $k=2$, while Section~\ref{sec:k_seq_attacks} generalizes such scenario to the ones in which the Attacker may perform an arbitrary number $k$ of attacks.
%

\subsection{Two Sequential Attacks} \label{sec:two_seq_attacks}
We can formalize the game with $k=2$ as follows. At the root of the game tree, the Defender chooses a vertex to place her resource. Then, the Attacker selects her action: she may attack two targets with a simultaneous attack, generating $|T|(|T|-1)$ nodes, or attack one target or wait, thus generating $|T|+1$ nodes. In the case the Attacker makes two simultaneous attacks or performs the second attack, the subgame can be solved by invoking \texttt{SolveSRG} (if the attacks are not simultaneous the penetration time of the first attacked target must be reduced by the number of turns the first attack is ongoing). Otherwise, \Def~has at most $|V|$ actions, corresponding to the vertices adjacent to $v_s$. For each of these actions, we have again $|T|(|T|-1)$ and $|T|+1$ nodes. The construction of the tree is performed iterating this process. Thus, the number of nodes at turn $\tau$ is $\sum_{l=1}^{\tau}|T|(|T|-1)|V|^l$. In principle, the game tree is infinite, making the computation of the equilibrium intractable. 

The fact that the attacks may be sequential makes covering routes unsuitable. In fact, a covering route should specify a set of targets protected by the patroller, but with sequential attacks, a covering route at a given turn may not be \textit{covering} after a further attack. 
We thus extend the concept of \emph{direct route}, generalizing it as follows.

\begin{definition}[Route] \label{def:route_general}
	Given $\mathsf{attacked}(0)$, a {\em route} is a couple $r=(p_r,T_r)$, where $p_r=(p_r(0), \ldots, p_r(h))$ is a sequence of arbitrary finite length $h$, with $p_r(i)$ being any vertex of $G$, and $T_r=(T_r(0),\ldots,T_r(j))$ is the set containing all the targets under attack reached by the patroller within their penetration times during her moving along $p_r$.
\end{definition}

Any other target $t$ not appearing in $T_r$ is not visited or visited after $d(t)$ turns from the start of the attack. When \Def~plays a route $r$, all and only the targets appearing in $T_r$ are protected. It would appear natural that the Defender still moves along the shortest paths to go from an attacked target to another one, thus minimizing the response time. Unfortunately, the following result holds.

\begin{proposition} \label{prop:not_sp}
	Moving along shortest paths between targets can be a dominated strategy for the Defender.
\end{proposition}

\begin{figure}[!htbp]
	\centering
	\begin{tabular}{c}
		\begin{tikzpicture}
		[scale=.6,auto=left,every node/.style={circle,fill=white,draw,minimum size = 0.675cm,font=\sffamily\large\bfseries}]
		\node (n1) at (0,0) {$v_2$};
		\node (n2) at (-4,0) {\Def};
		\node (n3) at (-2,0) {$v_1$};
		\node (n4) at (2,0) {$v_3$};
		\node (n5) at (4,0) {$t_1$};
		\node (n6) at (-2,2) {$v_5$};
		\node (n7) at (2,2) {$v_4$};
		\node (n8) at (0,-2) {$t_2$};
		
		\path[every node/.style={font=\sffamily\small}]
		(n1) edge node [loop] {} (n3)
		(n1) edge node [loop] {} (n4)
		(n2) edge node [loop] {} (n3)
		(n4) edge node [loop] {} (n5)
		(n2) edge node [loop] {} (n6)
		(n6) edge node [loop] {} (n7)
		(n7) edge node [loop] {} (n5)
		(n3) edge node [loop] {} (n8)
		(n1) edge node [loop] {} (n8)
		(n4) edge node [loop] {} (n8);
		\end{tikzpicture}
	\end{tabular}
	\vspace{-0.2cm}
	\caption{Graph to prove Proposition~\ref{prop:not_sp}.}
	\vspace{-0.2cm}
	\label{fig:not_sp}
\end{figure}
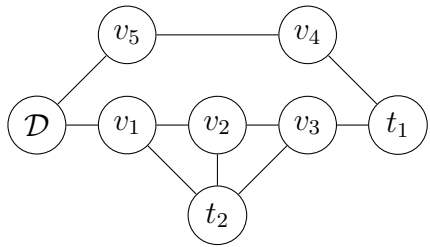

\noindent
\textit{Proof.}
The proof is by an example. Given the graph depicted in Figure~\ref{fig:not_sp}, let all the edges be unitary, the penetration times of the targets be equal to $d(t_1)=5,d(t_2)=3$, respectively, and the Attacker be able to perform at most two attacks. Moreover, let $v_\mathcal{D}$ be the starting position of the patroller and the first attack be performed against $t_1$. The shortest path to $t_1$ is $\langle v_\mathcal{D},v_5,v_4,t_1 \rangle$: however, while following this path, if the Attacker perform her second attack against $t_2$ while the patroller is in $v_5$, one target is necessarily lost.
Conversely, if the patroller moves along $\langle v_\mathcal{D},v_1,v_2,v_3,t_1 \rangle$, she can save both targets, independently from her current position. This concludes the proof.
\hfill $\Box$

To find the best strategy of the patroller, we provide the following exact algorithm, \texttt{PathFinder}.

\begin{algorithm}[!htbp]
	\caption{\texttt{PathFinder($v_s,t'$)}}
	\begin{algorithmic}[1]
		\State $\forall i,j \in V \times V, M(i,j) = (r_{si},r^c_{ij},u_{ij}) \leftarrow (\langle \cdot\rangle,\langle \cdot\rangle,0)$\label{loc:init_m}
		\State $M(s,1) = (\langle v_s \rangle,\langle \cdot \rangle,0)$\label{loc:init_el}
		\ForAll{$j,i \in V \times V$ s.t. $M(i,j) \neq (\langle \cdot\rangle,\langle \cdot\rangle,0)$}\label{loc:init_iter}
		\State $(r^c_{min},u_{min}) \leftarrow$ \texttt{AttackPrediction($v_i,t',j$)}\label{loc:call_att_pred}
		\State $M(i,j) \leftarrow (\cdot,r^c_{min},\min(u_{min},u_{ij}))$ \label{loc:update}
		\ForAll{$(v_i,v_{adj}) \in E$} \label{loc:init_update_adj}
		\If{$u_{ij} \leq u_{adj,j+1}$}
		\State $M(adj,j+1)\leftarrow(\langle r_{si},v_{adj}\rangle,\cdot,u_{ij})$
		\EndIf
		\EndFor \label{loc:end_update_adj}
		\EndFor\label{loc:end_iter}
		\State $u^* \leftarrow \max(u_{1|V|},u_{|V||V|})$ \label{loc:best_u}
		\State \textbf{return} $(r^*,r^{c,*},u^*) \leftarrow M(i,j) | M(i,j)=(\cdot,\cdot,u^*)$ \label{loc:return}
	\end{algorithmic}
	\label{alg:two_seq_att}
\end{algorithm}

\begin{algorithm}[!htbp]
	\caption{\texttt{AttackPrediction($v,t',j$)}}
	\begin{algorithmic}[1]
		\State $d(t')\leftarrow d(t')-j$
		\ForAll{$t \in T, t \neq t'$}\label{loc_att:init_iter}
		\State $(R,U) \leftarrow$ \texttt{SolveSRG}$(v,\{t\}\cup\{t'\})$ \label{loc_att:solve}
		\EndFor \label{loc_att:end_iter}
		\State \textbf{return} $(r^c_{min},u_{min}) \leftarrow (\argmin \limits_i U_i,\min \limits_i U_i)$ \label{loc_att:return}
	\end{algorithmic}
	\label{alg:att_pred}
\end{algorithm}

Let the first attack be performed against target $t'$ with the patroller in $v_s$. Our goal is to figure out if we can cover $t'$ within its penetration time and, if so, what route should be followed. \texttt{PathFinder} builds the paths from $v_s$ to $t'$ incrementally, assuming that the second attack may be performed at each step towards $t'$, and solving this problem by invoking the \texttt{AttackPrediction} algorithm. 
\texttt{PathFinder} uses a $|V| \times |V|$ matrix $M$, where each element $M(i,j) = (r_{si},r^c_{ij},u_{ij})$ consists of the route $r_{si}$ from $v_s$ to $v_i$, the best covering route $r^c_{ij}$ from vertex $v_i$ at time instant $j$ to cover the targets under attack and the utility $u_{ij}$ for \Def~associated to such route. 
The rows represent the vertices of the graph while the columns represent the time steps. Each cell is initialized to $(\langle \cdot\rangle,\langle \cdot\rangle,0)$ except for the one corresponding to the starting point, $M(v_s,1)$, which contains the route constituted only by $v_s$ (Algorithm~\ref{alg:two_seq_att}, Lines~\ref{loc:init_m}-\ref{loc:init_el}).
$M$ is filled column by column, as time passes by (Algorithm~\ref{alg:two_seq_att}, Lines~\ref{loc:init_iter}-\ref{loc:end_iter}).
\texttt{PathFinder} processes elements of $M$ corresponding to visited vertices. Let us focus on a cell of the matrix, $M(i,j)$.
\texttt{AttackPrediction} is called on $v_i$ (Algorithm~\ref{alg:two_seq_att}, Line~\ref{loc:call_att_pred}): first, it reduces the penetration time of $t'$ according to the turn we are considering, then it calls \texttt{SolveSRG} to obtain the set of covering routes $R$ that should be followed if the second attack would happen against $t \neq t'$ with \Def~in $v_i$ at time instant $j$ and the utilities $U$ associated  to the routes (Algorithm~\ref{alg:att_pred}, Lines~\ref{loc_att:init_iter}-\ref{loc_att:end_iter}).
After performing such computation, since \Att~will carry on her second attack in our worst-case scenario, the algorithm saves the covering route and the value corresponding to the worst attack in $r^c_{min},u_{min}$, respectively, (Algorithm~\ref{alg:att_pred}, Line~\ref{loc_att:return}) and return them back.
The values of the current cell are updated, taking $r^c_{min}$ and the minimum value between $u_{min}$ and the previous one contained in the cell, namely $u_{ij}$ (Algorithm~\ref{alg:two_seq_att}, Line~\ref{loc:update}).
The values of the cells corresponding to vertices $v_{adj}$ adjacent to $v_i$ are also updated: if the minimum value between the previous one contained in $M(adj,j+1)$ and the updated value of $u_{ij}$ is the latter, the route from $v_s$ to $v_i$ to which $v_{adj}$ is appended and $u_{ij}$ are saved (Algorithm~\ref{alg:two_seq_att}, Lines~\ref{loc:init_update_adj}-\ref{loc:end_update_adj}).
Finally, the algorithm returns the element of $M$ containing the highest utility for the Defender (Algorithm~\ref{alg:two_seq_att}, Line~\ref{loc:return}), i.e., either the best path to reach $t'$ or standing still in $v_s$.

\begin{theorem}
	\emph{\texttt{PathFinder}} computes the equilibrium path of the game, returning the optimal solution for the Defender.
\end{theorem}

\noindent
\textit{Proof.} We report the proof, based on the following lemmas.

\begin{lemma} \label{lemma:tree_size}
	At the equilibrium, the size of the game tree is finite.
\end{lemma}

\noindent
\textit{Proof.}
From Lemma~\ref{lemma:not_twice}, we know that the patroller will never visit twice the same node and so the path of the Defender cannot be longer than $|V|$. Moreover, Lemma~\ref{lemma:attack_one_way} tell us that the second attack will be performed while the patroller is traveling to cover the first target. Thus, the depth of the game tree is limited by $|V|$.
\hfill $\Box$

\begin{lemma}\label{lemma:attack_one_way} 
	Let the first attack being performed against target $t'$ with the patroller in $v_s$. Then, the Attacker will perform the second attack while the patroller is moving from $v_s$ to~$t'$.
\end{lemma}

\noindent
\textit{Proof.}
We consider the second attack occurring when the Defender is going from $v_s$ to $t'$ along $p$ and on her way back, from $t'$ to $v_s$, following path $p'$ (following a path different from $p$ on the way back is a dominated action). If the Attacker can complete the attack while \Def~is moving along $p'$, then the same attack can also be completed when she is moving along $p$. But if the Attacker cannot complete the attack while \Def~is going back to $v_s$, then \Att~might be able to complete the attack when \Def~is traveling along $p$ according to the patrolling policy.$\Box$

\begin{lemma} \label{lemma:not_twice}
	Let the first attack be performed against target $t'$ with the patroller in $v_s$. Then, in absence of the second attack, the Defender will never traverse twice the same vertex along her path from $v_s$ to $t'$.
\end{lemma}

\noindent
\textit{Proof.}
Let $p$ be the path from $v_s$ to $t'$ followed by \Def~when no other attack is occurring. Being the graph unitary, the patroller will reach $t'$ after $p$ time units. If \Def~should traverse a vertex twice, she would reach $t'$ in more than $p$ time units, without getting any increase in terms of utility. Thus, \Def~will follow $p$. 
\hfill $\Box$

\begin{lemma}\label{lemma:path_dom} 
	Let $p_1$, $p_2$ be two paths that visit the same node $v$ at time instant $\tau$ and $u_1,u_2$ the utilities obtained considering the worst-attack that can be performed when traveling along such paths. If $u_1 \geq u_2$, then traveling along $p_1$ dominates traveling along $p_2$.
\end{lemma}

\noindent
\textit{Proof.}
After visiting $v$, both paths will have the same expected utility associated to the next steps towards $t'$. Thus, we can compare $p_1,p_2$ with respect to  their utilities $u_1',u_2'$ in reaching $v$. Since $u_1 \geq u_2$, then $u_1' \geq u_2'$ and so traveling along $p_1$ dominates traveling along $p_2$.
\hfill $\Box$

Because of Lemmas~\ref{lemma:tree_size}--\ref{lemma:path_dom}, we can state that \texttt{PathFinder} evaluates all the solutions among which the optimal ones may be, safely discarding only dominated ones. 
Thus, eventually, the algorithm will return the optimal solution, and the corresponding strategy, for the Defender.
This concludes the proof sketch.
\hfill $\Box$

Before running \texttt{PathFinder}, we compute the shortest paths among all the vertices, with a cost of $O(|V^3|)$ (Floyd-Warshall algorithm~\cite{cormen2009introduction}). 
We build a matrix $M$, whose size is $|V|^2$, and the cost of filling each $M(i,j)$ is dominated by the calls to \texttt{AttackPrediction}, whose cost is $|V|$ (invoking \texttt{SolveSRG} has a cost that is constant in $k$ since it is a fixed parameter). The complexity of invoking \texttt{PathFinder} is then $O(|V|^3)$.
We call \texttt{PathFinder} for each $t' \in T$, returning the best routes with the associated utilities. At the end of these executions, we know which are the best responses for the patroller from $v_s$ when any target is attacked.
We repeat this procedure for each vertex $v \in V$, reaching a total complexity of $O(|V|^5)$.
After this, the Defender knows, for each $v_i \in V$, the utility $u_{v_i}^*$ of placing the patrolling resource in $v_i$ and how to best respond from there to each possible attack. The best starting placement for the patroller is then $v^* = \argmax (u_{v_1}^*,u_{v_2}^*,\ldots,u_{v_{|V|}}^*)$.

Thus, once $k$ is fixed, the computation of the equilibrium path can be done in polynomial time. Moreover, if the Attacker is irrational, i.e., playing off the equilibrium path, \texttt{PathFinder} can still be used to find the best response in polynomial time.

\subsection{Extending to an Arbitrary Number of Sequential Attacks} \label{sec:k_seq_attacks}
The extension of \texttt{PathFinder} to an arbitrary number of resources is involved: in order to introduce the new features, we first investigate what happens when the Attacker performs the first attack with $k-1$ resources, and then we apply the proposed approach to the most general case.

Let us analyze what happens if \Att~performs her first attack employing $k-1$ resources. We introduce two additional features with respect to  \texttt{PathFinder}. First, we add a third dimension to $M$, say $l$, considering all the combinations with repetitions of the $k-1$ targets under attack (a target that has already been attacked but not successfully compromised can be attacked again). We move along $l$ according to the currently active targets, thus excluding targets covered by the patroller or those that have been successfully attacked by the Attacker. Moreover, we need to explicitly keep track of the targets that have actually been covered by the patroller. Indeed, since there could be multiple attacks, the Defender may traverse some target after it has been successfully attacked, but we should not include it among the targets that have been successfully covered. Thus, each element of $M$ also contains  the set of covered targets.

We follow steps similar to \texttt{PathFinder}, filling $M$ column by column and for increasing $l$, calling an extended version of \texttt{AttackPrediction} that takes as input the subset of targets $T'$ under attack. According to the targets that have been successfully attacked by the Attacker and the ones covered by the covering route, we fill the corresponding cell of $M$ by inserting covered targets, the corresponding route and value, and then we update the elements of the cells of the adjacent vertices by adopting the same rationale employed by \texttt{PathFinder}. Once $M$ has been completely processed, we select the last matrix along the $l$ dimension and apply the same operations we performed at the end of \texttt{PathFinder} (Algorithm~\ref{alg:two_seq_att}, Lines~\ref{loc:best_u}-\ref{loc:return}), returning the element of $M$ associated with the best utility for the Defender.

Here, the size of $M$ is $|V|\cdot (k-1)|V|\cdot \binom{|T|+k-1}{k}$ and this time calling \texttt{AttackPrediction} has a cost equal to $O(2^kk^5)$ due to the call to \texttt{SolveSRG}. So, the cost of invoking the extended version of \texttt{PathFinder} is $O((|V|+k)^k2^kk^6|V|^2)$.
We run this algorithm for all subsets of $k-1$ targets, namely $\binom{|T|}{k-1}$, to know which are the best responses for the patroller from $v_s$ when any subset of $k-1$ targets is attacked.
We then repeat this procedure for each $v \in V$ and select the vertex with the highest utility as the starting placement for the patroller, reaching a total complexity of $O((|V|+k)^k|V|^{k+2}2^kk^6)$ to solve our problem.

Now let us consider the general case in which \Att~employs $k-k'$ resources for the first attack. This means we must solve all the problems with $k-k'+1$ possible targets under attack and for each of these, all the problems with $k-k'+2$. We proceed recursively until we reach problems with $k-1$ targets under attack: we solve these problems as described above and propagate back the solutions, solving step by step all the problems until we reach the original one.

To compute the solution for the general problem, we have to solve a very large number of problem, namely $\prod_{h=1}^{k'-1} \binom{|T|}{k-k'+h}\approx O(|V|^{k^2})$, each with the exponential complexity showed above.

\section{Without Complete Information} \label{sec:no_prior}
Till now, we assumed that the Defender \textit{a priori} knew the number of resources the Attacker could employ to perform her attack. 
In real-life scenarios, however, it is very unlikely that this information is available for the Defender to be used. A possible approach to face this issue is employing a probability distribution on the number of resources available to the Attacker, as required in Bayesian games. 
Unfortunately, in this case, the problem is even computationally harder, and the assumption that such a prior is common knowledge is even more unlikely. Thus, we study the scenario in which knowledge about the number of attacker's resources is not common.
First, Section~\ref{sec:robustness} analyzes the robustness when the Defender makes a wrong guess on the number of resources actually controlled by the Attacker, while Section~\ref{sec:online_algorithms} proposes two online algorithms to deal with this problem when the Defender has no information about the number of the resources.

\subsection{Robustness to a Wrong Guess} \label{sec:robustness}
In this section,  we investigate the scenario in which \Def~makes a guess $k'$ about the number of resources available to \Att, being such a number actually equal to $k$. 
First, as customary done in the literature, we adopt a ratio to evaluate the quality of the guess: in this case, we consider the ratio between the Defender's utilities obtained with the wrong guess and the correct guess as our measure~\ref{sec:ratio}. 
Then, we resort to another index of performance for the guess, namely, the difference of the Defender's utilities.

\subsubsection{Utility Ratio} \label{sec:ratio}
Initially, we study the loss in the worst-case for the Defender, i.e., she plays her best strategy against $k'$ resources when such guess is wrong. Moreover, we assume the Attacker to be rational and knowing the guess $k'$ made by the Defender.
We denote with $\sigma^*_{\mathcal{D},k}$ and with $v^*_{\mathcal{D},k}$, respectively, the optimal strategy of the patroller and the value of the equilibrium when the number of resources available to \Att~is $k$ and the guess is correct. Similarly, $v_{\mathcal{D},k,k'}$ is the value obtained by the Defender playing strategy $\sigma_{\mathcal{D},k,k'}$ when the Attacker has $k$ resources but the guess is $k'$, and \Att~plays her best response to $\sigma_{\mathcal{D},k,k'}$.

We define the relative loss of a strategy based on a wrong guess with respect to the best strategy based on the correct guess by resorting to the tools used in online algorithms, where the performance of an algorithm is compared with respect to  the performance of the clairvoyant algorithm, which \emph{a priori} knows all the information. In particular, we resort to the concept of \emph{competitive factor}~\cite{fiat1998online}.

\begin{definition}[Competitive factor]
	The competitive factor $\Gamma$ of an algorithm is given by the worst-case ratio $\frac{v}{v^*}$, where $v$ is the value given by the algorithm and $v^*$ is the maximum value given by a clairvoyant algorithm.
\end{definition}

We recall that an algorithm is said \emph{competitive} when $\Gamma>0$. In our case, the competitive factor when the Defender makes a guess~$k'$ while the Attacker has $k$ resources is equal to $\Gamma=\frac{v_{\mathcal{D},k,k'}}{v^*_{\mathcal{D},k}}$, once excluded all the instances in which $v^*_{\mathcal{D},k} = 0$. 

Before tackling the problem, we have to normalize the values of the targets onto the interval $[0,1]$ to have correct values for the competitive factor. 
Let $T_{topk}$ be a list of targets obtained re-labeling the targets in $T$ in descending order  with respect to  $\pi(t)$ and selecting the first $k$.
Given an instance where $\pi(t)$ are the initial values for the targets, we change their values as follows: $\pi'(t)=\frac{\pi(t)}{\sum_{t_i \in T_{topk}} \pi(t_i)}$.
This way, the Defender gets a utility equal to $1$ if no targets are conquered by the Attacker while she gets $0$ if all the top $k$ targets are conquered. In other words, a utility of $0$ means that no other outcome is worse for the Defender.

Now we can state the following.

\begin{theorem}\label{thm:bad_guess}
	If the Defender makes a wrong guess with respect to  the actual number of resources available to the Attacker, then $\Gamma=0$, independently by the value of the guess.
\end{theorem}

\noindent
\textit{Proof.}
We consider separately the case in which the Defender is \emph{underestimating} the number of the attacker's resources from the case in which she is \emph{overestimating} it. 

\textit{Underestimation}. Let us consider a graph composed of a clique of size $k-k'+1$, with unitary edges, $k-k'$ targets, say $t_c$, with $\pi(t_c)=1,d(t_c)=k-k'$, and the non-target vertex $v$ of the clique connected to $k'$ other targets, $t_o$, with $\pi(t_o)=\epsilon,d(t_o)=k$, through edges with weight equal to $k$. If $k'=k$, starting in $v$, \Def~will lose at most $(k-1)\epsilon$ since she will wait for attacks to targets $t_c$ and, if they happen, the patroller will be able to cover all of them. According to our normalization, $v^*_{\mathcal{D},k}=1$. On the other hand, if $k'<k$, \Att~will employ $k'$ resources on targets $t_o$ and $k-k'$ to attack $t_c$. From $v$, \Def~will be able to save only a target $t_o$, so $v_{\mathcal{D},k,k'}=\epsilon$. Since $\Gamma=\epsilon$, $\Gamma\xrightarrow{\epsilon\rightarrow0}0$.

\textit{Overestimation}. Let us consider a \textit{star} graph with unitary edges and $k$ targets connected to a central node $v$. The values of the targets are $\pi(t)=1-\epsilon$ except for one target, $t_{max}$, whose value is $\pi(t_{max})=1$. From $v$, if \Def~could employ the clairvoyant algorithm, she would always be able to save one target, thus $v^*_{\mathcal{D},k}=1-\frac{(k-1)(1-\epsilon)}{1+(k-1)(1-\epsilon)}=\frac{1}{1+(k-1)(1-\epsilon)}$. 
On the other side, since \Att~employs $k<k'$ resources, the Defender will wait for the last $k-k'$ targets to be attacked since $t_{max}$ could be among them. In this case, $v_{\mathcal{D},k,k'}=1-\frac{k(1-\epsilon)}{1+(k-1)(1-\epsilon)}=\frac{\epsilon}{1+(k-1)(1-\epsilon)}$. Thus, $\Gamma\xrightarrow{\epsilon\rightarrow0}0$.
\hfill $\Box$

The above results shows that playing a strategy that is optimal for a given guess is not robust in practice since it is sufficient that the guess is wrong just by one resource to have an arbitrary loss with respect to  the optimal clairvoyant solution.

\subsubsection{Utility Difference}\label{sec:difference}
As shown in the previous section, considering just the ratio of the utilities may not be very informative since making an error when performing a guess can lead to an arbitrary loss.

This is why we turn our attention to another index that could be more significant for our setting, i.e., the difference between the Defender's utilities computed when w.r.t the guessed number of resources and the correct number of resources adopted by the Attacker, respectively. 
Moreover, adopting such a criterion is also suggested by the concept of the $\epsilon$-Nash equilibrium~\cite{shoham2008multiagent}, which is commonly adopted when dealing with robustness analysis of games subject to perturbations.

\begin{definition}[Additive competitive factor]
	The additive competitive factor $\Gamma'$ of an algorithm is given by the worst-case difference $v - v^*$, where $v$ is the value given by the algorithm and $v^*$ is the maximum value given by a clairvoyant algorithm.
\end{definition}

We observe that $\Gamma'$ has been defined such that, if the guess is wrong, it results negative, meaning the Defender is incurring in some loss. In this case, adopting a similar approach, it can proved that the loss strictly depends on the difference between the guess $k'$ of the Defender and the actual number of resources available to the Attacker, i.e., $k$, as stated in the following.

\begin{theorem}
	If the Defender makes a wrong guess with respect to  the actual number of resources available to the Attacker, then $\Gamma'$ is equal to:
	\begin{center}
		$\Gamma' = \begin{cases} -(k-k') + \epsilon, & k'<k \\
		-(k'-k)(1-\epsilon), & k<k'<2k \\
		-k(1-\epsilon), & k' \geq 2k.
		\end{cases}$
	\end{center}
	where $k'$ is the number of resources guessed by the Defender while $k$ is the actual number of resources controlled by the Attacker.
\end{theorem}

\noindent
\textit{Proof.}
We consider separately the case in which the Defender is \emph{underestimating} the number of the attacker's resources from the case in which she is \emph{overestimating} it. 

\textit{Underestimation}. 
Let us consider a graph composed of a clique of size $k-k'+1$, with unitary edges, $k-k'$ targets, say $t_c$, with $\pi(t_c)=1,d(t_c)=k
$, and the non-target vertex $v$ of the clique connected to $k'$ other targets, $t_o$, with $\pi(t_o)=\epsilon,d(t_o)=k$, through edges with weight equal to $k$. 
If $k'=k$, starting in $v$, \Def~will lose at most $k'\epsilon$ due to attacks on targets $t_o$ since she will wait for attacks to targets $t_c$ and, if they happen, the patroller will be able to cover all of them. Thus, $v^*_{\mathcal{D},k}=-k'\epsilon$.
Conversely, if $k'<k$, \Att~will employ $k'$ resources on targets $t_o$ and $k-k'$ to attack $t_c$. From $v$, \Def~will be able to save only a target $t_o$, so $v_{\mathcal{D},k,k'}=-k-(k'-1)\epsilon$. 
Thus, $\Gamma' = -k-(k'-1)\epsilon - k'\epsilon = -(k-k') + \epsilon$.

\textit{Overestimation}. 
We consider two different cases, first evaluating what happens if $k < k' \geq 2k$ and then considering the case $k'>2k$.

\textit{$k < k' \geq 2k$}. Let us consider a graph composed of a clique of size $k'-k+1$, with unitary edges, $k'-k$ targets, say $t_c$, with $\pi(t_c)=1,d(t_c)=k'-k$, and the non-target vertex $v$ of the clique connected to $k$ other targets $t_o$ through edges with weight equal to $k$, with such targets characterized by $\pi(t_o)=1-\epsilon,d(t_o)=k'-k$. Targets $t_o$ are connected to each other by edges with weight equal to 1, forming a clique.
If $k'=k$, it can be seen that, because of the structure of the instance, the best strategy of the Attacker is to perform half of the attacks against targets $t_c$ and the other half against targets $t_o$. The worst case is when $k$ is odd, inflicting the Defender a loss equal to $(2k-k')(1-\epsilon)$, since she will prefer to save the targets $t_c$ with respect to  $t_o$ since protecting the former guarantees her a lower loss. Thus, $v^*_{\mathcal{D},k}=-(2k-k')(1-\epsilon)$

If $k < k' \geq 2k$, \Att~will employ $k$ resources to attack targets on $t_o$. If the Defender moves to defend these targets, she knows the Attacker will perform the remaining $k'-k$ attacks against $t_c$ targets. This way, the Defender loss would be equal to $1+(k-1)(1-\epsilon)$, since she would be able to save only one $t_o$ target and $k-1$ $t_c$ targets, losing all the others.
Conversely, if \Def~stands still, despite losing all the $t_o$ targets, she would be able to save all the $t_c$ targets, and so her loss would be equal t $k(1-\epsilon)$.
Since standing still the loss of the Defender is smaller, she will adopt this strategy. Thus, $v_{\mathcal{D},k,k'}=-k(1-\epsilon)$
This means that $\Gamma' =  -k(1-\epsilon) + (2k-k')(1-\epsilon) = -(k'-k)(1-\epsilon)$.

\textit{$k'>2k$}. Let us consider a graph composed of a clique of size $k'-k+1$, with unitary edges, $k'-k$ targets, say $t_c$, with $\pi(t_c)=1,d(t_c)=k+1$, and the non-target vertex $v$ of the clique connected to $k$ other targets $t_o$ through edges with weight equal to $k$, with such targets characterized by $\pi(t_o)=1-\epsilon,d(t_o)=k+1$. Targets $t_o$ are connected to each other by edges with weight equal to 1, forming a clique.
If $k'=k$, it can be seen that, because of the structure of the instance, from $v$ the Defender will be able to save all the targets, independently of the sequence with which they are performed. Thus, $v^*_{\mathcal{D},k}=0$.
If $k'>2k$, \Att~will employ $k$ resources to attack targets on $t_o$. If the Defender moves to defend these targets, she knows the Attacker will perform the remaining $k'-k$ attacks against $t_c$ targets. This way, the Defender loss would be equal to $1+(k-1)(1-\epsilon)$, since she would be able to save only one $t_o$ target and $k-1$ $t_c$ targets, losing all the others.
Conversely, if \Def~stands still, despite losing all the $t_o$ targets, she would be able to save all the $t_c$ targets, and so her loss would be equal to $k(1-\epsilon)$.
Since standing still the loss of the Defender is smaller, she will adopt this strategy. Thus, $v_{\mathcal{D},k,k'}=-k(1-\epsilon)$.
This means that $\Gamma' = -k(1-\epsilon) - 0 = -k(1-\epsilon)$.

\noindent
This concludes the proof.
\hfill $\Box$

\begin{observation}
The above values for $\Gamma'$ are tight.
	\begin{itemize}
		\item Underestimation: $\Gamma'\xrightarrow{\epsilon\rightarrow0}-(k-k')$. This is the highest value the Attacker can get since she must employ $k'$ resources to deceive the Defender, who otherwise will wait for a possible attack against a high-valuable target.
		\item Overestimation, $k < k' \geq 2k$: $\Gamma'\xrightarrow{\epsilon\rightarrow0}-(k'-k)$. This is the highest value the Attacker can get since, being the guess strictly smaller than $2k$, she must employ some resources to deceive the Defender, who will move to block the resources and save some targets.
		\item Overestimation, $k'>2k$:  $\Gamma'\xrightarrow{\epsilon\rightarrow0}-k$. This is the biggest loss the Attacker can inflict to the Defender with $k$ resources.
	\end{itemize}

Moreover, notice that when $k<k'\geq 2k$, the loss increase of one unit as the guess is bigger than one unit with respect to  $k$. This holds until we reach $k'=2k$, where the loss is equal to $k$. Here, the loss obtained in the two overestimating cases smoothly connect, since for $k'>2k$ the loss is constantly equal to $k$.
\end{observation}

\subsection{Online Algorithms} \label{sec:online_algorithms}
We have seen that if the guess of the Defender on the number of resources available to the Attacker is wrong, underestimating or, more surprisingly, overestimating such number leads to an arbitrarily small value for \Def. Thus, we ask whether there exists a competitive online algorithm that is independent of the actual number of resources of the Attacker, just relying only on the observed attacks, so that can circumvent the problem.

\begin{theorem}\label{thm:online_det}
	There is no deterministic online competitive algorithm with a competitive factor better than $\frac{1}{k-1}$, where $k$ is the actual number of resources available to the Attacker.
\end{theorem}

\noindent
\textit{Proof.}
Let us consider the following instance. A vertex $v$ is connected to $\max\{1,k-1\}$ targets, with deadlines equal to $\max\{1,2k-1\}$ through unitary edges, and to $t_f$ with an edge whose cost is $2k$, where $d(t_f)=2k$. The value of all the targets is $\pi(t)=1$.

We split the proof considering first $k=1$ and then $k>1$. If $k=1$, with the patroller in $v$, the optimal strategy for \Def~is to protect the first target under attack, obtaining a value of $1$. Any online algorithm that does not prescribe to cover the first target under attack will have a competitive factor equal to $0$ since the Defender will take $0$.
If $k>1$, the optimal strategy for the Defender is to cover the first target under attack if and only if such target is not $t_f$. This way, in the optimal case, the Defender will take a utility of $\frac{k-1}{k}$, protecting all the targets except for one. Conversely, any strategy prescribing to defend $t_f$ when attacked as the first target would lead to a competitive factor of $\frac{1}{k-1}$. Thus, the best competitive factor is $\frac{1}{k-1}$.
\hfill $\Box$

It is now worth asking whether we can achieve a better result employing randomization\footnote{In this work, we just open the path to the study of this problem, aware that this a fundamental question we will further investigate in future works.}.

\begin{theorem}
	Let $\Gamma_d$ be the best competitive factor of a deterministic online algorithm. There exists a randomized online algorithm with competitive factor $\Gamma_r$ such that $\Gamma_r > \Gamma_d$ and, asymptotically, $\Gamma_r \rightarrow \Gamma_d$.
\end{theorem}

\noindent
\textit{Proof.}
We prove the theorem only in the worst case for deterministic online algorithms. This allows us to show the improvement one can obtain by means of randomization in the worst case.

Let us consider a graph composed of a clique with unitary edges of $k-h+1$ nodes, $k-h$ of which are targets, say $t_c$, with $d(t_c)=k-h$, and the non-target vertex $v$ of the clique connected to $h$ other targets, $t_o$, with $d(t_o)=k$, through edges with weight equal to $k$. $\pi(t_c)=\pi(t_o)=1$.

We know from Theorem~\ref{thm:online_det} that a deterministic approach reaches a competitive factor $\Gamma_d=\frac{1}{k-1}$. We propose the following randomized algorithm: let $v$ be the starting point for \Def~and, whenever an attack occurs, she has probability $\frac{1}{2}$ of protecting the target under attack while  she stands still in $v$ with probability $\frac{1}{2}$. Since we want to compute $\Gamma_r$, we normalize the values of the targets dividing them by $k$. If there are multiple sequential attacks on targets $t_o$, the utility of protecting them is equal to $\frac{1}{k}\left(\frac{1}{2}+\frac{1}{4}+\cdots+\frac{1}{2^h}\right)=\frac{1}{2k}\sum_{i=0}^h\left(\frac{1}{2}\right)^i$ while standing still and covering attacks against $t_c$ gives a utility equal to $\left(\frac{1}{2}\right)^h\frac{k-h}{k}$. The clairvoyant algorithm achieves a utility of $\frac{k-h}{k}$. Thus, $\Gamma_r=\frac{\frac{1}{2k}\sum_{i=0}^h\left(\frac{1}{2}\right)^i+\left(\frac{1}{2}\right)^h\frac{k-h}{k}}{\frac{k-h}{k}}=\frac{1}{k-h}\left(1-\left(\frac{1}{2}\right)^{h+1}\right)+\left(\frac{1}{2}\right)^h$. Among all the instances, we want to find the worst, so, given $k$, we want to minimize $\Gamma_r$ with respect to  $h$. In the following table, we report the values of $\Gamma_r$ and $\Gamma_d$ achieved for different $k$, taking the $h$ minimizing $\Gamma_r$.

\begin{small}
	\begin{center}
		\begin{tabular}{|c|c|c|c|c|c|c|c|c|c|}
			\hline
			$k$ & 3 & 4 & 5 & 6 & 7 & 8 & 9 & 10 & 100\\ \hline
			$\Gamma_r$ & 0.87 & 0.69 & 0.54 & 0.44 & 0.36 & 0.30 & 0.26 & 0.22 & 0.01\\ \hline
			$\Gamma_d$ & 0.50 & 0.33 & 0.25 & 0.20 & 0.17 & 0.14 & 0.12 & 0.11& 0.01\\
			\hline
		\end{tabular}\label{table:rand_vs_det}
	\end{center}
\end{small}

As it can be seen, for small values, a very simple randomized algorithm can ensure a competitive factor that is twice better than the one achieved by a deterministic algorithm. Moreover, $\Gamma_r \xrightarrow{k\rightarrow \infty} \Gamma_d$.
\hfill $\Box$

\section{Conclusions and Future Research} \label{sec:conclusions}
In this work, we investigated the opportunities an Attacker can take when she can perform multiple attacks, simultaneously or sequentially, in an arbitrary environment, modeled as a graph. 
The challenge is due to the high interaction level among the players, e.g., the Attacker can use resources to make the patroller move away from some valuable targets and, subsequently, attack those targets.
Since the problem presents an explicit representation of the passing of time, we modeled it as an extensive-form game. 
In principle, an equilibrium can be found in polynomial time in the size of the game tree, but, here, the game tree induced by our model is exponentially large in the size of the graph and in the number of resources available to the Attacker. 
When the number of resources is a fixed parameter, the problem admits an algorithm capable of finding the strategies \emph{on the equilibrium path} requiring polynomial time in the size of the graph.
Conversely, we show that there is no algorithm requiring polynomial time in the number of Attacker's resources, unless $\mathsf{P}=\mathsf{NP}$, even in the simplified case in which the Attacker uses all her resources simultaneously. 
Unfortunately, computing the equilibrium strategies requires the knowledge on the number of Attacker's resources. 
Since it is unlikely to have this information, we studied the robustness of a Defender's strategy when the guess about the number of resources the Attacker can employ is wrong. 
We evaluated the worst-case inefficiency of this strategy showing that it can be arbitrary even when the guess is a wrong estimate---both over and under---for just a single resource.
We also investigated the problem looking at an additive competitive factor, according to an $\epsilon$-Nash equilibrium fashion.
Furthermore, we investigated the use of \emph{online} algorithms to adopt when no information is available to the Defender. 
We provided a tight upper bound over the \emph{competitive factor} when non-stochastic online algorithms are used, and we show that the factor can be improved by resorting to randomization.

The work presented in this paper can be extended along different directions.
We could enrich our model with respect to  the uncertainties that characterize the alarm system, introducing false positives. Even though an Attacker with multiple resources could actually recreate a similar effect, performing an attack just to deceive the Defender and then attacking her main target, this is not the same as having the system affected by such an issue. In fact, the Attacker could exploit this flaw, while the Defender should decide whether it is convenient or not to move from her current position. 
Similarly, we could add the presence of missed detections, i.e., even though an attack is occurring, no signal is raised by the system. This drawback affects all commercial alarm system, and thus it would be another important step towards a more realistic model.
Finally, we could deepen the impact an Attacker can have with various resources, being able to damage the targets at different levels.
This would lead to consider different levels of damage for the targets, giving the Attacker the possibility to stop an ongoing attack once a certain damage threshold is reached.

\bibliographystyle{plain}
\bibliography{refs}

\begin{thebibliography}{10}

\bibitem{adler2003pursue}
Micah Adler, Harald R\"{a}cke, Naveen Sivadasan, Christian Sohler, and Berthold
  V\"{o}cking.
\newblock {Randomized Pursuit-Evasion in Graphs}.
\newblock {\em Combinatorics, Probability and Computing}, 12:225--244, 2003.

\bibitem{agmon2010events}
Noa Agmon.
\newblock {On Events in Multi-robot Patrol in Adversarial Environments}.
\newblock In {\em International Conference on Autonomous Agents and Multi-Agent
  Systems {(AAMAS)}}, pages 591--598, 2010.

\bibitem{agmon2012coordination}
Noa Agmon, Chien-Liang Fok, Yehuda Emaliah, Peter Stone, Christine Julien, and
  Sriram Vishwanath.
\newblock {On Coordination in Practical Multi-robot Patrol}.
\newblock In {\em {IEEE} International Conference on Robotics and Automation
  {(ICRA)}}, pages 650--656, 2012.

\bibitem{agmon2011multi}
Noa Agmon, Gal~A. Kaminka, and Sarit Kraus.
\newblock {Multi-robot Adversarial Patrolling: Facing a Full-knowledge
  Opponent}.
\newblock {\em Journal of Artificial Intelligence Research}, 42:887--916, 2011.

\bibitem{alpern1992inf}
Steve Alpern.
\newblock {Infiltration Games on Arbitrary Graphs}.
\newblock {\em Journal of Mathematical Analysis and Applications},
  163:286--288, 1992.

\bibitem{alpern2011patrolling}
Steve Alpern, Alec Morton, and Katerina Papadaki.
\newblock {Patrolling Games}.
\newblock {\em Operations Research}, 59(5):1246--1257, 2011.

\bibitem{amigoni2009finding}
Francesco Amigoni, Nicola Basilico, and Nicola Gatti.
\newblock {Finding the Optimal Strategies for Robotic Patrolling with
  Adversaries in Topologically-represented Environments}.
\newblock In {\em {IEEE} International Conference on Robotics and Automation
  {(ICRA)}}, pages 819--824, 2009.

\bibitem{amigoni2010moving}
Francesco Amigoni, Nicola Basilico, Nicola Gatti, Alessandro Saporiti, and
  Stefano Troiani.
\newblock {Moving Game Theoretical Patrolling Strategies from Theory to
  Practice: An USARSim Simulation}.
\newblock In {\em {IEEE} International Conference on Robotics and Automation
  {(ICRA)}}, pages 426--431, 2010.

\bibitem{an2012protect}
Bo~An, Eric Shieh, Milind Tambe, Rong Yang, Craig Baldwin, Joseph DiRenzo, Ben
  Maule, and Garrett Meyer.
\newblock {PROTECT - A Deployed Game Theoretic System for Strategic Security
  Allocation for the United States Coast Guard}.
\newblock {\em AI Magazine}, 33(4):96, 2012.

\bibitem{basilico2017coordinating}
Nicola Basilico, Andrea Celli, Giuseppe De~Nittis, and Nicola Gatti.
\newblock {Coordinating Multiple Defensive Resources in Patrolling Games with
  Alarm Systems}.
\newblock In {\em International Conference on Autonomous Agents and Multi-Agent
  Systems {(AAMAS)}}, pages 678--686, 2017.

\bibitem{basilico2016security}
Nicola Basilico, Giuseppe De~Nittis, and Nicola Gatti.
\newblock {A Security Game Combining Patrolling and Alarm-triggered Responses
  under Spatial and Detection Uncertainties}.
\newblock In {\em Conference on Artificial Intelligence {(AAAI)}}, pages
  404--410, 2016.

\bibitem{basilico2017adversarial}
Nicola Basilico, Giuseppe De~Nittis, and Nicola Gatti.
\newblock {Adversarial Patrolling with Spatially Uncertain Alarm Signals}.
\newblock {\em Artificial Intelligence}, 246:220--257, 2017.

\bibitem{basilico2011automated}
Nicola Basilico and Nicola Gatti.
\newblock {Automated Abstractions for Patrolling Security Games}.
\newblock In {\em Conference on Artificial Intelligence {(AAAI)}}, 2011.

\bibitem{basilico2012patrolling}
Nicola Basilico, Nicola Gatti, and Francesco Amigoni.
\newblock {Patrolling Security Games: Definition and Algorithms for Solving
  Large Instances with Single Patroller and Single Intruder}.
\newblock {\em Artificial Intelligence}, 184:78--123, 2012.

\bibitem{basilico2009capturing}
Nicola Basilico, Nicola Gatti, and Thomas Rossi.
\newblock {Capturing Augmented Sensing Capabilities and Intrusion Delay in
  Patrolling-Intrusion Games}.
\newblock In {\em {IEEE} Symposium on Computational Intelligence and Games
  {(CIG)}}, pages 186--193, 2009.

\bibitem{basilico2010asynchronous}
Nicola Basilico, Nicola Gatti, and Federico Villa.
\newblock {Asynchronous Multi-robot Patrolling against Intrusions in Arbitrary
  Topologies}.
\newblock In {\em Conference on Artificial Intelligence {(AAAI)}}, 2010.

\bibitem{birkhoff1946tres}
Garrett Birkhoff.
\newblock {Tres Observaciones sobre el Algebra Lineal}.
\newblock {\em Universidad Nacional de Tucum{\'a}n. Facultad de Ciencias
  Exactas y Tecnolog{\'i}a. Revista. Serie A. Matem{\'a}tica y F{\'i}sica
  Te{\'o}rica}, 5:147--151, 1946.

\bibitem{brown2016one}
Matthew Brown, Arunesh Sinha, Aaron Schlenker, and Milind Tambe.
\newblock {One Size Does Not Fit All: A Game-Theoretic Approach for Dynamically
  and Effectively Screening for Threats}.
\newblock In {\em Conference on Artificial Intelligence {(AAAI)}}, pages
  425--431, 2016.

\bibitem{carpin2007usarsim}
Stefano Carpin, Mike Lewis, Jijun Wang, Stephen Balakirsky, and Chris Scrapper.
\newblock {USARSim: A Robot Simulator for Research and Education}.
\newblock In {\em {IEEE} International Conference on Robotics and Automation
  {(ICRA)}}, pages 1400--1405, 2007.

\bibitem{chatterjee2012survey}
Krishnendu Chatterjee and Thomas~A. Henzinger.
\newblock {A Survey of Stochastic $\omega$-regular Games}.
\newblock {\em Journal of Computer and System Sciences}, 78(2):394--413, 2012.

\bibitem{conitzer2006computing}
Vincent Conitzer and Tuomas Sandholm.
\newblock {Computing the Optimal Strategy to Commit to}.
\newblock In {\em ACM conference on Electronic Commerce {(EC)}}, pages 82--90,
  2006.

\bibitem{cormen2009introduction}
Thomas~H. Cormen.
\newblock {\em {Introduction to Algorithms}}.
\newblock MIT Press, 2009.

\bibitem{delle2014game}
Francesco~Maria Delle~Fave, Albert~Xin Jiang, Zhengyu Yin, Chao Zhang, Milind
  Tambe, Sarit Kraus, and John~P. Sullivan.
\newblock {Game-theoretic Patrolling with Dynamic Execution Uncertainty and a
  Case Study on a Real Transit System}.
\newblock {\em Journal of Artificial Intelligence Research}, 50:321--367, 2014.

\bibitem{durkota2015approximate}
Karel Durkota, Viliam Lis{\`y}, Branislav Bo{\v{s}}ansk{\`y}, and Christopher
  Kiekintveld.
\newblock {Approximate Solutions for Attack Graph Games with Imperfect
  Information}.
\newblock In {\em International Conference on Decision and Game Theory for
  Security {(GameSec)}}, pages 228--249, 2015.

\bibitem{durkota2015optimal}
Karel Durkota, Viliam Lis{\`y}, Branislav Bosansk{\`y}, and Christopher
  Kiekintveld.
\newblock {Optimal Network Security Hardening Using Attack Graph Games}.
\newblock In {\em International Joint Conference on Artificial Intelligence
  {(IJCAI)}}, pages 526--532, 2015.

\bibitem{fang2015security}
Fei Fang, Peter Stone, and Milind Tambe.
\newblock {When Security Games Go Green: Designing Defender Strategies to
  Prevent Poaching and Illegal Fishing}.
\newblock In {\em International Joint Conference on Artificial Intelligence
  {(IJCAI)}}, pages 2589--2595, 2015.

\bibitem{fiat1998online}
Amos Fiat.
\newblock {\em {Online Algorithms: The State of the Art}}.
\newblock Springer, 1998.

\bibitem{flood1972hide}
Merrill~M. Flood.
\newblock {The Hide and Seek Game of Von Neumann}.
\newblock {\em Management Science}, 18(5-part-2):107--109, 1972.

\bibitem{ford2016nectar}
Benjamin Ford, Amulya Yadav, Amandeep Singh, Matthew Brown, Arunesh Sinha,
  Biplav Srivastava, Christopher Kiekintveld, Nicole Sintov, and Milind Tambe.
\newblock {NECTAR: Game-Theoretic Factory Inspection Scheduling and Explanation
  for Toxic Wastewater Abatement}.
\newblock In {\em International Conference on Autonomous Agents and Multi-agent
  Systems {(AAMAS)}}, 2016.

\bibitem{gal1980search}
Shmuel Gal.
\newblock {\em {Search Games}}.
\newblock Academic Press, 1980.

\bibitem{gan2015security}
Jiarui Gan, Bo~An, and Yevgeniy Vorobeychik.
\newblock {Security Games with Protection Externalities}.
\newblock In {\em Conference on Artificial Intelligence {(AAAI)}}, pages
  914--920, 2015.

\bibitem{gan2017security}
Jiarui Gan, Bo~An, Yevgeniy Vorobeychik, and Brian Gauch.
\newblock {Security Games on a Plane}.
\newblock In {\em Conference on Artificial Intelligence {(AAAI)}}, pages
  530--536, 2017.

\bibitem{gholami2016divide}
Shahrzad Gholami, Bryan Wilder, Matthew Brown, Dana Thomas, Nicole Sintov, and
  Milind Tambe.
\newblock {Divide to Defend: Collusive Security Games}.
\newblock In {\em International Conference on Decision and Game Theory for
  Security {(GameSec)}}, pages 272--293, 2016.

\bibitem{kar2017trends}
Debarun Kar, Thanh~H Nguyen, Fei Fang, Matthew Brown, Arunesh Sinha, Milind
  Tambe, and Albert~Xin Jiang.
\newblock {Trends and Applications in Stackelberg Security Games}.
\newblock {\em Handbook of Dynamic Game Theory}, pages 1--47, 2017.

\bibitem{kiekintveld2009computing}
Christopher Kiekintveld, Manish Jain, Jason Tsai, James Pita, Fernando
  Ord{\'o}{\~n}ez, and Milind Tambe.
\newblock {Computing Optimal Randomized Resource Allocations for Massive
  Security Games}.
\newblock In {\em International Conference on Autonomous Agents and Multi-Agent
  Systems {(AAMAS)}}, pages 689--696, 2009.

\bibitem{korzhyk2010complexity}
Dmytro Korzhyk, Vincent Conitzer, and Ronald Parr.
\newblock {Complexity of Computing Optimal Stackelberg Strategies in Security
  Resource Allocation Games}.
\newblock In {\em Conference on Artificial Intelligence {(AAAI)}}, 2010.

\bibitem{munoz2013introducing}
Enrique Munoz~de Cote, Ruben Stranders, Nicola Basilico, Nicola Gatti, and Nick
  Jennings.
\newblock {Introducing Alarms in Adversarial Patrolling Games}.
\newblock In {\em International conference on Autonomous Agents and Multi-Agent
  Systems {(AAMAS)}}, pages 1275--1276, 2013.

\bibitem{nguyen2016capture}
Thanh~H Nguyen, Arunesh Sinha, Shahrzad Gholami, Andrew Plumptre, Lucas Joppa,
  Milind Tambe, Margaret Driciru, Fred Wanyama, Aggrey Rwetsiba, Rob Critchlow,
  et~al.
\newblock {Capture: A New Predictive Anti-poaching Tool for Wildlife
  Protection}.
\newblock In {\em International Conference on Autonomous Agents and Multi-agent
  Systems}, pages 767--775, 2016.

\bibitem{papadaki2016patrolling}
Katerina Papadaki, Steve Alpern, Thomas Lidbetter, and Alec Morton.
\newblock Patrolling a border.
\newblock {\em Operations Research}, 64(6):1256--1269, 2016.

\bibitem{paruchuri2008playing}
Praveen Paruchuri, Jonathan~P. Pearce, Janusz Marecki, Milind Tambe, Fernando
  Ordonez, and Sarit Kraus.
\newblock {Playing Games for Security: An Efficient Exact Algorithm for Solving
  Bayesian Stackelberg Games}.
\newblock In {\em International Joint Conference on Autonomous Agents and
  Multi-Agent Systems {(AAMAS)}}, pages 895--902, 2008.

\bibitem{pita2008deployed}
James Pita, Manish Jain, Janusz Marecki, Fernando Ord\'{o}\~{n}ez, Christopher
  Portway, Milind Tambe, Craig Western, Praveen Paruchuri, and Sarit Kraus.
\newblock {Deployed ARMOR Protection: The Application of a Game-theoretic Model
  for Security at the Los Angeles International Airport}.
\newblock In {\em International Conference on Autonomous Agents and Multi-Agent
  Systems {(AAMAS)}}, pages 125--132, 2008.

\bibitem{pita2011guards}
James Pita, Milind Tambe, Chris Kiekintveld, Shane Cullen, and Erin
  Steigerwald.
\newblock {GUARDS: Game Theoretic Security Allocation on a National Scale}.
\newblock In {\em International Conference on Autonomous Agents and Multi-Agent
  Systems {(AAMAS)}}, pages 37--44, 2011.

\bibitem{ruckle1976ambush}
William Ruckle, Robert Fennell, Paul~T. Holmes, and Charles Fennemore.
\newblock {Ambushing Random Walks I: Finite Models}.
\newblock {\em Operations Research}, 24(2):314--324, 1976.

\bibitem{schlenker2016get}
Aaron Schlenker, Matthew Brown, Arunesh Sinha, Milind Tambe, and Ruta Mehta.
\newblock {Get Me to My GATE on Time: Efficiently Solving General-Sum Bayesian
  Threat Screening Games}.
\newblock In {\em European Conference on Artificial Intelligence {(ECAI)}},
  pages 1476--1484, 2016.

\bibitem{schlenker2017don}
Aaron Schlenker, Haifeng Xu, Mina Guirguis, Chris Kiekintveld, Arunesh Sinha,
  Milind Tambe, Solomon Sonya, Darryl Balderas, and Noah Dunstatter.
\newblock {Don't Bury your Head in Warnings: A Game-Theoretic Approach for
  Intelligent Allocation of Cyber-security Alerts}.
\newblock In {\em International Joint Conference on Artificial Intelligence
  {(IJCAI)}}, 2017.

\bibitem{shieh2013efficiently}
Eric Shieh, Manish Jain, Albert~Xin Jiang, and Milind Tambe.
\newblock {Efficiently Solving Joint Activity Based Security Games}.
\newblock In {\em International Joint Conference on Artificial Intelligence
  {(IJCAI)}}, pages 346--352, 2013.

\bibitem{shoham2008multiagent}
Yoav Shoham and Kevin Leyton-Brown.
\newblock {\em {Multiagent systems: Algorithmic, Game-theoretic, and Logical
  Foundations}}.
\newblock Cambridge University Press, 2008.

\bibitem{sless2014multi}
Efrat Sless, Noa Agmon, and Sarit Kraus.
\newblock {Multi-robot Adversarial Patrolling: Facing Coordinated Attacks}.
\newblock In {\em International Conference on Autonomous Agents and Multi-agent
  Systems {(AAMAS)}}, pages 1093--1100, 2014.

\bibitem{tsai2009iris}
J.~Tsai, S.~Rathi, C.~Kiekintveld, F.~Ord{\'{o}}{\~{n}}ez, and M.~Tambe.
\newblock {IRIS - A Tool for Strategic Security Allocation in Transportation
  Networks}.
\newblock In {\em International Conference on Autonomous Agents and Multi-Agent
  Systems {(AAMAS)}}, pages 1327--1334, 2009.

\bibitem{von2004leadership}
Bernhard Von~Stengel and Shmuel Zamir.
\newblock {Leadership with Commitment to Mixed Strategies}.
\newblock Technical report, 2004.

\bibitem{an2014adp}
Yevgeniy Vorobeychik, Bo~An, Milind Tambe, and Satinder~P. Singh.
\newblock {Computing Solutions in Infinite-Horizon Discounted Adversarial
  Patrolling Games}.
\newblock In {\em International Conference on Automated Planning and Scheduling
  {(ICAPS)}}, pages 314--322, 2014.

\bibitem{wang2017stop}
Xinrun Wang, Qingyu Guo, and Bo~An.
\newblock {Stop Nuclear Smuggling Through Efficient Container Inspection}.
\newblock In {\em International Conference on Autonomous Agents and Multi-Agent
  Systems {(AAMAS)}}, pages 669--677, 2017.

\bibitem{yin2016efficient}
Yue Yin and Bo~An.
\newblock {Efficient Resource Allocation for Protecting Coral Reef Ecosystems}.
\newblock In {\em International Conference on Artificial Intelligence
  {(IJCAI)}}, pages 531--537, 2016.

\bibitem{zhang2017optimal}
Youzhi Zhang, Bo~An, Long Tran-Thanh, Zhen Wang, Jiarui Gan, and Nicholas~R.
  Jennings.
\newblock {Optimal Escape Interdiction on Transportation Networks}.
\newblock In {\em International Conference on Artificial Intelligence
  {(IJCAI)}}, pages 3936--3944, 2017.

\bibitem{zhao2016optimizing}
Mengchen Zhao, Bo~An, and Christopher Kiekintveld.
\newblock {Optimizing Personalized Email Filtering Thresholds to Mitigate
  Sequential Spear Phishing Attacks}.
\newblock In {\em Conference on Artificial Intelligence {(AAAI)}}, pages
  658--665, 2016.

\end{thebibliography}

\newpage
\appendix
\section{Notation Table} \label{appendix:notation}
We report in Table~\ref{tab:activityTracking} the symbols used throughout the paper.

\begin{table}[!h]
	\centering
	\begin{tabular}{|l|l|l|}
		\cline{2-3}
		\multicolumn{1}{ c| }{} & Symbol & Meaning \\
		\cline{2-3} \hline
		\multirow{11}{*}{\rotatebox{90}{Basic model}}
		& $\mathcal{A}$ & Attacker \\ \cline{2-3}
		& $\mathcal{D}$ & Defender \\ \cline{2-3}
		& $G=(V,E)$ & Graph constituted by the set of vertices $V$ and the set of edges $E$\\ \cline{2-3}
		& $v$ & Vertex \\ \cline{2-3}
		& $(v, v')$ & Edge \\ \cline{2-3}
		& $\omega^*_{v, v'}$ & Temporal cost (in turns) of the shortest path between $v$ and $v'$ \\ \cline{2-3}
		& $T$ & Set of targets \\ \cline{2-3}
		& $t$ & Target \\ \cline{2-3}
		& $t_i$ & $i$-th target \\ \cline{2-3}    
		& $\pi(t)$ & Value of target $t$ \\ \cline{2-3}
		& $d(t)$ & Penetration time of target $t$ \\
		
		\hline
		
		\multirow{6}{*}{\rotatebox{90}{Signals}}
		& $S$ & Set of signals \\ \cline{2-3}
		& $s_i$ & Signal associated to target $t_i$ \\ \cline{2-3}
		& $p$ & Function specifying the probability of having the system generating \\
		& & signal $s$ given that target t has been attacked \\ \cline{2-3}
		& $T(s)$ & Targets having a positive probability of raising $s$ if attacked \\ \cline{2-3}
		& $S(t)$ & Signals having a positive probability of being raised if $t$ is attacked \\
		
		\hline   
		
		\multirow{11}{*}{\rotatebox{90}{Actions, routes, strategies}}
		& $\tau$ & Turn of the game \\ \cline{2-3}
		& $\mathsf{attacked}(\tau)$ & Set of targets attacked at tutn $\tau$ \\ \cline{2-3}
		& $(\tau,\mathsf{attacked}(\tau))$ & Action for the Attacker \\ \cline{2-3}
		& $k$ & Number of resources available to the Attacker \\ \cline{2-3}
		& $r$ & Route, i.e., sequence of potentially non-adjacent vertices \\ \cline{2-3}
		& $r_i$ & $i$-th route \\ \cline{2-3}
		& $r(i)$ & $i$-th element visited along route $r$ \\ \cline{2-3}
		& $R$ & Set of routes \\ \cline{2-3} 
		& $A(r(i))$ & Time needed by $\mathcal{D}$ to visit $r(i)$ starting from $r(0)$ \\ \cline{2-3}
		& $T(r)$ & Set of targets covered by route $r$ \\ \cline{2-3}
		& $c(r)$ & Temporal cost (in turns) associated to $r$ \\ 
		
		\hline
		
		\multirow{7}{*}{\rotatebox{90}{Online analysis}} 
		& $\Gamma$  & Multiplicative competitive factor\\ \cline{2-3}     
		& $\Gamma'$  & Additive competitive factor\\ \cline{2-3}
		& $\sigma^*_{\mathcal{D},k}$ & Optimal Defender's strategy against $k$ attacks\\ \cline{2-3}
		& $\sigma^*_{\mathcal{D},k,k'}$ & Defender's strategy when she guesses $k'$ attacks \\ \cline{2-3}
		& $v^*_{\mathcal{D},k}$ & Optimal Defender's utility against $k$ attacks\\ \cline{2-3}
		& $v_{\mathcal{D},k,k'}$ & Defender's utility when she guesses $k'$ attacks \\ 
		& & and \Att perform $k$ attacks \\
		
		\hline
	\end{tabular}
	\caption[Symbols' table]{Symbols' table.}
	\label{tab:activityTracking}
\end{table}

\end{document}